\documentclass[dvipsnames]{article}

\usepackage[numbers]{natbib}
\setcitestyle{numbers,square,comma}
\usepackage[final]{lmrm}
\usepackage[utf8]{inputenc} 
\usepackage[T1]{fontenc}    
\usepackage{url}            
\usepackage{booktabs}       
\usepackage{amsfonts}       
\usepackage{nicefrac}       
\usepackage{booktabs} 
\usepackage{multirow}
\usepackage{wrapfig}
\usepackage{amssymb}
\usepackage{epigraph}
\usepackage{makecell}
\usepackage{listings}

\usepackage[labelfont=bf]{caption} 
\usepackage[ruled,vlined]{algorithm2e}
\usepackage{color, soul}
\usepackage{microtype}      
\usepackage{xcolor}         
\usepackage{geometry}
\usepackage{times}
\usepackage{ragged2e}
\usepackage{amsmath}
\usepackage{bm}
\usepackage{latexsym}
\usepackage{subfigure}
\usepackage{graphicx}
\usepackage{float}
\usepackage{longtable}
\usepackage{booktabs} 
\usepackage{multirow}
\usepackage{amssymb}
\usepackage{longtable}
\usepackage{tabu}
\usepackage{bbding}
\usepackage{awesomebox}
\usepackage{bbding}
\usepackage[most,breakable]{tcolorbox}
\usepackage{etoolbox}
\usepackage{lipsum}
\usepackage{CJKutf8}
\usepackage{pdfpages}
\usepackage{multicol}
\usepackage{pgffor} 
\usepackage{fontawesome5}
\usepackage{nicematrix} 
\usepackage{paralist}
\usepackage[edges]{forest}
\usepackage{siunitx}
\usepackage{tikz-dependency}
\usepackage{tikz}
\usepackage{bbm}
\usepackage{amssymb}
\usepackage[english]{babel}  
\usepackage{hyphenat}
\usepackage{enumitem}
\usepackage{pifont}

\newcommand{\Rmnum}[1]{\expandafter\@slowromancap\romannumeral #1@}

\usepackage{caption} 
\usepackage{chngcntr}

\usepackage[colorlinks, linkcolor=RoyalBlue, anchorcolor=BrickRed, citecolor=RoyalBlue, urlcolor=RoyalBlue]{hyperref}

\usepackage{cleveref}
\crefname{section}{§}{§§}
\Crefname{section}{§}{§§}

\newcommand\refsec[1]{Section~\hyperref[sec:#1]{\ref{sec:#1}}}
\newcommand\refsecs[2]{\hyperref[sec:#1]{§\ref{sec:#1}:~\textsc{#1}}, \hyperref[sec:#2]{§\ref{sec:#2}:~\textsc{#2}}}

\usepackage[tikz]{bclogo}
\definecolor{msftBlue}{RGB}{0,164,239}
\definecolor{msftGreen}{RGB}{127,186,0}
\definecolor{msftYello}{RGB}{255,185,0}
\definecolor{myblue}{RGB}{30,136,229} 
\definecolor{msftBlack}{RGB}{0,0,0}
\definecolor{darkgreen}{RGB}{0,100,0}

\newtcolorbox{myboxnote}[1][]{
  breakable,
  title=#1,
  colback=cyan!0,
  colbacktitle=cyan!0,
  coltitle=black,
  fonttitle=\bfseries,
  bottomrule=0pt,
  toprule=0pt,
  leftrule=1.5pt,
  rightrule=1.5pt,
  titlerule=0pt,
  arc=0pt,
  outer arc=0pt,
  colframe=lightgray,
}
\usepackage{mdframed}

\definecolor{academicblue}{RGB}{54, 95, 145}

\newtcbox{\smybox}[1][red]{on line,
arc=1pt,colback=#1!10!white,colframe=#1!100!black,
before upper={\rule[-3pt]{0pt}{10pt}},
boxsep=0pt,left=6pt,right=6pt,top=2pt,bottom=0pt,boxrule=0pt,leftrule=1pt,rightrule=1pt}

\definecolor{mybluebg}{HTML}{f3f8fb}    
\definecolor{myclueborder}{HTML}{6faad0} 
\definecolor{mybluetitle}{HTML}{bcd7e9}  
\definecolor{myshadow}{HTML}{d8d8d8}
\tcbset{
  iclrtakeawaybox/.style={
    width=\linewidth,
    colback=mybluebg,                
    colframe=myclueborder,          
    colbacktitle=mybluetitle,       
    coltitle=black,
    fonttitle=\bfseries,
    boxrule=0.5pt,
    arc=2pt,
    sharp corners=south,
    attach boxed title to top left={yshift=-0.1in,xshift=0.15in},
    boxed title style={boxrule=0pt, colframe=white},
    enhanced,
    drop shadow={
      shadow scale=1,
      shadow xshift=2pt,
      shadow yshift=-2pt,
      opacity=0.5,
      color=myshadow
    }
  }
}
\newtcolorbox{TakeawayBox}[2][]{iclrtakeawaybox,title=#2,#1}

\newenvironment{itemsize*}%
 {\leftmargini=20pt\begin{itemize}%
  \setlength{\itemsep}{3pt}%
  \setlength{\parskip}{0pt}%
  }%
 {\end{itemize}}
\newenvironment{enumerate*}%
 {\begin{enumerate}%
  \setlength{\itemsep}{0pt}%
  \setlength{\parskip}{0pt}}%
 {\end{enumerate}}

\usepackage{blindtext} 
\usepackage[normalem]{ulem}

\usepackage{xparse}
\usepackage{colortbl}

\newcommand{\benchmark}{\textsc{VideoMathQA}}

\definecolor{small}{RGB}{255, 255, 255} 
\definecolor{big}{RGB}{148, 153, 192} 
\newcommand{\rgbrank}[2]{
  \pgfmathsetmacro{\percent}{%
    \ifnum#2<10
      0 + #2
    \else
      \ifnum#2<30
        10 + (#2-10) * 0.5
      \else
        \ifnum#2<45
          20 + (#2-30) * 1.33
        \else
          50 + (#2-45) * 0.91
        \fi
      \fi
    \fi
  }%
  \edef\temp{\noexpand\cellcolor{big!\percent!small}}\temp
  \temp #1
}

\title{VideoMathQA: Benchmarking Mathematical Reasoning \\ via Multimodal Understanding in Videos}

\author{
\large Hanoona Rasheed$^{1}$, Abdelrahman Shaker$^{1}$, Anqi Tang$^{1}$, Muhammad Maaz$^{1}$\\
\large \textbf{Ming-Hsuan Yang$^{2,3}$, Salman Khan$^{1,4}$, Fahad Shahbaz Khan$^{1,5}$} \\[0.5ex]
\normalsize $^{1}$MBZUAI \quad
$^{2}$University of California Merced \quad
$^{3}$Google Research \\
\normalsize $^{4}$Australian National University \quad
$^{5}$Linköping University \\
\normalsize \textcolor{blue}{\url{https://mbzuai-oryx.github.io/VideoMathQA}}
}

\begin{document}

\maketitle

\begin{abstract}
Mathematical reasoning in real-world video settings presents a fundamentally different challenge than in static images or text. It requires interpreting fine-grained visual information, accurately reading handwritten or digital text, and integrating spoken cues, often dispersed non-linearly over time. In such multimodal contexts, success hinges not just on perception, but on selectively identifying and integrating the right contextual details from a rich and noisy stream of content. To this end, we introduce \benchmark, a benchmark designed to evaluate whether models can perform such temporally extended cross-modal reasoning on videos. 
The benchmark spans 10 diverse mathematical domains, covering videos ranging from 10 seconds to over 1 hour. It requires models to interpret structured visual content, understand instructional narratives, and jointly ground concepts across visual, audio, and textual modalities. We employ graduate-level experts to ensure high quality, totaling over $920$ man-hours of annotation.
To reflect real-world scenarios, questions are designed around three core reasoning challenges: \emph{direct problem solving}, where answers are grounded in the presented question; \emph{conceptual transfer}, which requires applying learned methods to new problems; and \emph{deep instructional comprehension}, involving multi-step reasoning over extended explanations and partially worked-out solutions. 
Each question includes multi-step reasoning annotations, enabling fine-grained diagnosis of model capabilities. Through this benchmark, we highlight the limitations of existing approaches and establish a systematic evaluation framework for models that must reason, rather than merely perceive, across temporally extended and modality-rich mathematical problem settings. 
Our benchmark and evaluation code are available at: \href{https://mbzuai-oryx.github.io/VideoMathQA}{VideoMathQA}.
\end{abstract}

\begin{figure}[htbp]
    \centering
    \begin{minipage}{0.82\linewidth}
        \includegraphics[width=\linewidth]{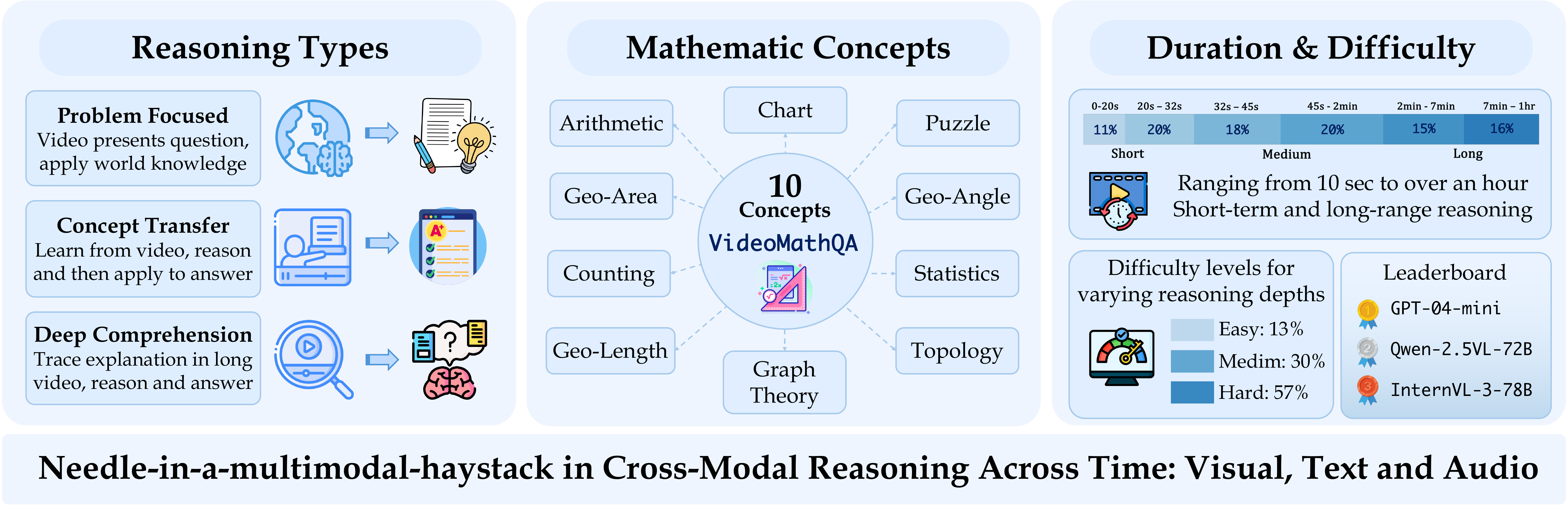}
        \caption{The foundation of our benchmark is the “needle-in-a-multimodal-haystack” challenge, capturing the core difficulty of cross-modal reasoning across time from visual, textual, and audio streams. Built on this, VideoMathQA categorizes each question along four key dimensions: reasoning type, mathematical concept, video duration, and difficulty.}
        \label{fig:enter-label}
    \end{minipage}
\end{figure}
\section{Introduction}
As opposed to mathematical reasoning benchmarks developed for static-image, video settings presents a set of unique challenges. In educational and instructional videos, key information is conveyed through evolving diagrams, handwritten or digital notations, and spoken explanations that unfold non‐linearly across time. A reasoning model must therefore sift through high-resolution frames, align visual representations with subtitles or voice‐over, and integrate disparate cues into a coherent problem‐solving pipeline. This \textit{`needle-in-a-multimodal-haystack'} problem requires not only accurate perception (e.g., frame-aware OCR of equations) but also precision in symbolic manipulation and multi-step inference, where missing a single visual or verbal cue can lead to incorrect conclusions.

\begin{figure*}[h]
  \centering
  \includegraphics[width=\textwidth]{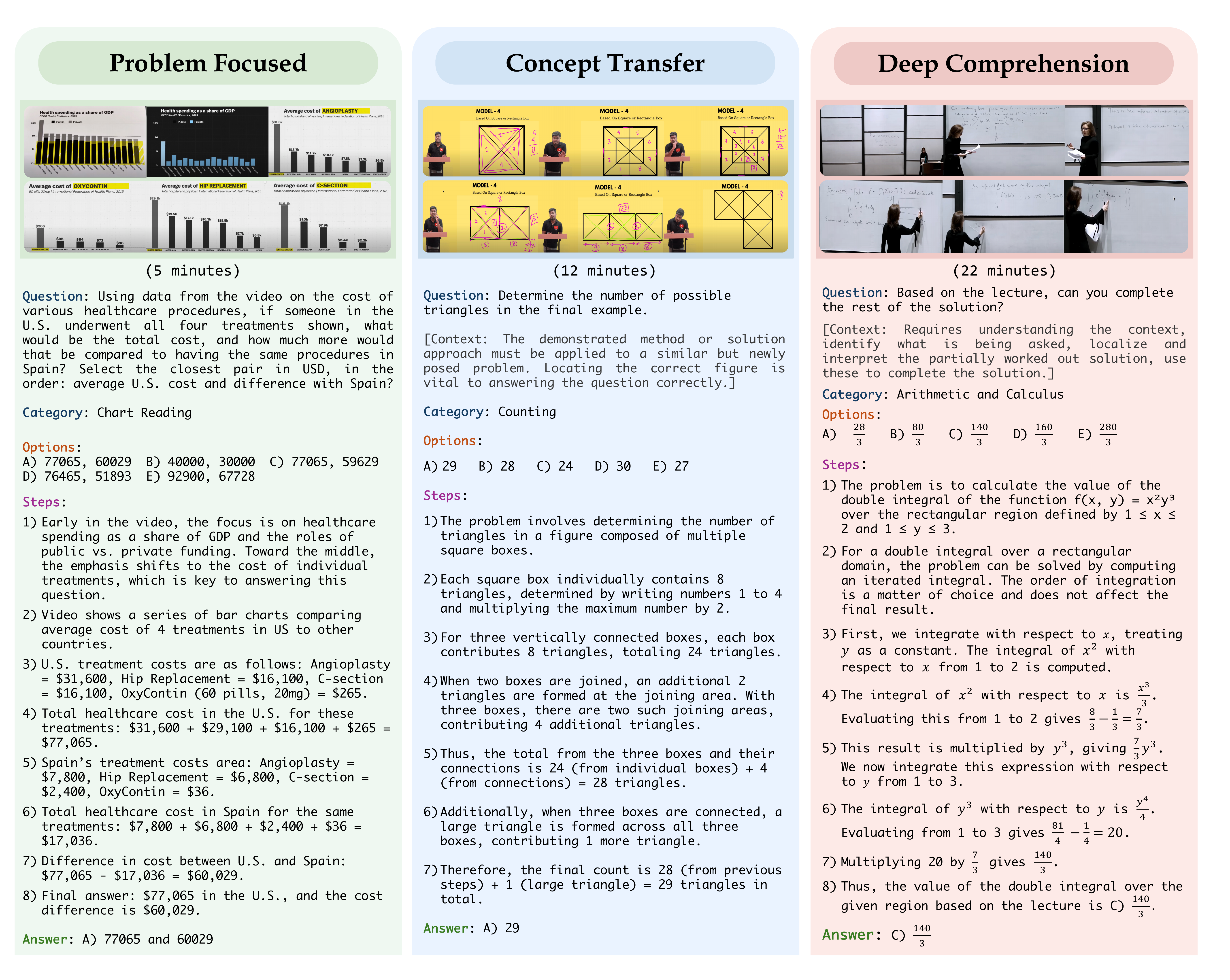}
  \caption{Example questions from the \benchmark~benchmark illustrating the three reasoning types: Problem Focused, Concept Transfer, and Deep Comprehension. The benchmark includes evolving dynamics in a video, complex text prompts, five multiple-choice options, the expert-annotated step-by-step reasoning to solve the given problem, and the final correct answer as shown above. }
  \label{fig:intro_fig}
\end{figure*}

Existing benchmarks for mathematical reasoning (e.g., MathQA~\cite{mathqa}, ChartQA~\cite{chartqa}, and MathVista~\cite{mathvista}) have driven substantial progress in image-based and text-based settings. However, they typically evaluate models on static diagrams, printed formulas, or single-turn queries and lack support for temporally extended contexts, dynamically evolving visual content, and do not offer true integration of vision, audio, text and background subject knowledge. Some recent efforts explore video question answering in general domains~\cite{videomme, egoschema, li2024mvbench}, but they do not target the precise, multi‐modal, multi‐step inference demanded by mathematical problem solving. Moreover, existing video benchmarks often rely on synthetic \cite{yi2019clevrer} or narrowly scoped tasks \cite{shangguan2024tomato} without detailed reasoning annotations, making it difficult to diagnose whether a model's solution stems from genuine logical inference or simple pattern matching.

To bridge these gaps, we introduce \benchmark, a comprehensive benchmark designed explicitly to evaluate deep mathematical reasoning in videos.
It comprises $420$ manually annotated real-world video–question pairs drawn from various educational resources.
\benchmark~tests three core reasoning scenarios:
\textit{a) Direct Problem Solving}, where answers are grounded entirely in the presented content;
\textit{b) Conceptual Transfer}, requiring agents to apply learned methods to novel problems; and
\textit{c) Deep Instructional Comprehension}, involving multi-step reasoning over extended, partially worked-out explanations.
Each instance includes high-resolution frames, aligned subtitles, and audio narration, and is annotated with fine-grained reasoning traces that capture every inference step to enable assessment of both intermediate steps and final outcomes.
Videos span ten mathematical domains (e.g., geometry, calculus, statistics, graph theory, chart) and range from $10$s clips to \emph{hour-long} lectures, ensuring both short-term perception and long-range dependency evaluation (see Fig.~\ref{fig:intro_fig} for illustrative examples of the reasoning scenarios). Our key contributions are:
\begin{itemize} 
    \item A multimodal video reasoning benchmark that demands precise integration of vision (high-resolution), text, audio and subject knowledge in solving complex math problems.
    \item Our benchmark offers three real-world task categories: problem solving, concept transfer, and deep comprehension, which reflect the spectrum of instructional scenarios in videos. It covers ten mathematical domains and various environment types (e.g., whiteboard scribbles, digital slides, animated charts), sourced from diverse sources.
    \item Fine-grained reasoning annotations and metrics for measuring both intermediate inference fidelity and final answer correctness, with explicit mechanisms to detect confabulations.
    \item A comprehensive evaluation across $30$ proprietary and open-source multimodal models. We also conduct a comprehensive analysis of model performance trends and failure modes.
\end{itemize}	

\begin{TakeawayBox}{Takeaways: Comprehensive Evaluation}
A comprehensive evaluation across $30$ models reveals that success relies not just on visual perception, but on sustained attention to subtle cues dispersed across time, modalities, and context. Models often fail when key frames, symbols, or spoken details are missed, revealing limited ability to integrate long-range multimodal information. While performance generally scales with size, architecture and training quality are often more decisive; newer, smaller models frequently outperform older, larger ones. Notably, the gap between proprietary and open-source systems is narrowing, as recent open models now match or exceed proprietary models.
\end{TakeawayBox}
\section{Related Works}
\label{sec:related_work}
\textbf{Video Multimodal Understanding Benchmarks.} Recent advances in multimodal understanding~\cite{gpt4o,liu2023visual,zhu2023minigpt} have led to the development of multiple video benchmarks. Early datasets like ActivityNet-QA~\cite{activitynetqa} and Next-QA~\cite{nextqa} focus on short clips and temporal action reasoning, while EgoSchema~\cite{egoschema} and MovieChat~\cite{moviechat} extend to long-form narrative comprehension. Recent efforts such as WorldQA\cite{worldqa}, Next-GQA\cite{Next-GQA}, VCGBench\cite{maaz2023video}, MVBench~\cite{li2024mvbench} and Video-MME~\cite{videomme} broaden the scope with diverse reasoning tasks. While these benchmarks have advanced multimodal understanding in everyday contexts, they fall short of the deeper complexity posed by mathematical reasoning, where models must navigate non-linear content and demonstrate tightly integrated understanding across vision, audio, text, and background subject knowledge.

\textbf{Multimodal Mathematical Benchmarks.} Recent image-based mathematical benchmarks have significantly advanced the evaluation of multimodal models. MathVista~\cite{mathvista} and Math-V~\cite{mathv} include visual questions drawn from textbooks and competitions, while MMMU~\cite{mmmu} and MMMU-Pro~\cite{mmmu_pro} introduce subject-specific questions with CoT prompts and OCR inputs. DynaMath~\cite{dynamath} evaluates the robustness of mathematical reasoning through visual perturbations. Although these benchmarks span a diverse range of mathematical topics, they are fundamentally limited to assessing reasoning over static images, and the temporal dimension intrinsic to video, where mathematical information may unfold through lectures, stepwise derivations, or interactive explanations, is not captured.

Video-MMMU~\cite{videommmu} begins to explore video-based academic QA, but includes only a small subset of mathematical questions and focuses on general comprehension. It does not target the depth of reasoning or modality integration that mathematical problem solving demands. In contrast, \benchmark~is designed to evaluate deep mathematical reasoning by challenging models to interpret high-resolution visuals, follow non-linear spoken explanations, and integrate vision, language, and domain knowledge over time, while providing detailed step-by-step reasoning annotations to enable comprehensive analysis beyond just final answer accuracy.
\section{\benchmark}

\subsection{Overview}
\label{sec:VIDEOMATHQA}
\begin{figure*}[t]
  \centering
  \includegraphics[width=\textwidth]{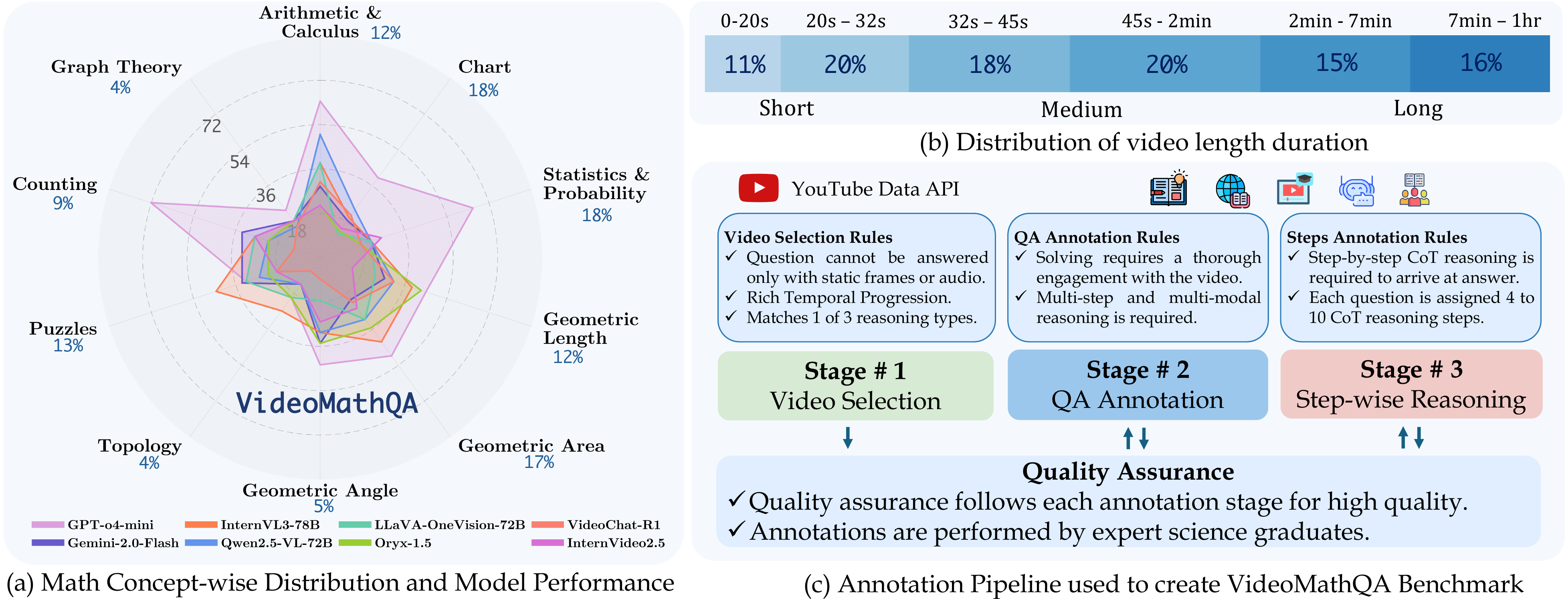}
  \caption{The figure illustrates \textbf{a)}~Distribution of questions and model performance across ten mathematical concepts in the \benchmark. The consistently low performance across all concepts reveals a significant gap in the ability of the current multimodal models to perform mathematical reasoning over videos. \textbf{b)}~Distribution of video durations in \benchmark, highlighting a diverse range from short clips of $10$s to long-videos up to $1$hr. \textbf{c)}~The three-stage annotation pipeline for \benchmark~was performed by expert science graduates, who annotated detailed step-by-step reasoning trails, with each stage governed by strict quality assessment.}
  \label{fig:overview}
  \vspace{-1.2em}
\end{figure*}

Despite the significant progress in image-based mathematical reasoning benchmarks, existing datasets fail to capture the unique challenges posed by video-based math problems. In designing \benchmark, we asked: \textit{How can we reliably evaluate mathematical reasoning in video-based settings, and what core challenges a benchmark must capture to reflect real-world multimodal understanding?} Our answers shaped the core principles of the benchmark. We identified \textit{three critical challenges}: the need to interpret \textbf{\textit{dynamic visual representations}}, such as diagrams constructed or modified over time, often involving handwritten or digital content requiring robust, frame-aware OCR; the need for \textbf{\textit{temporal reasoning over long, non-linear contexts}}, where concepts unfold gradually or are revisited across time; and the need for \textbf{\textit{joint grounding across visuals, text, and audio}}, as all three modalities often contribute distinct pieces of essential information. These interconnected challenges define the complexity of video-based mathematical tasks and directly inform the design of our benchmark.

We introduce \benchmark, a dataset of $420$ carefully curated video-question pairs drawn from diverse mathematical instructional content, including structured problem walkthroughs, concept demonstrations with follow-up questions, full-length whiteboard or digital lectures, and animated documentaries involving chart-based reasoning. Each question includes \textit{\textbf{multi-step reasoning steps}}, with a total of \textit{\textbf{2,945 expert-annotated steps}} across the dataset. Each question is characterized along \textit{four dimensions}: (\textit{\textbf{i}})~\textit{\textbf{Mathematical concept}}, covering $10$ domains such as geometry, arithmetic and calculus, statistics and probability, counting, graph theory, puzzles, topology, and chart reading,  (See Fig.~\ref{fig:overview}a); (\textbf{\textit{ii}})~\textit{\textbf{Reasoning type}}, categorized as problem focused, concept transfer, or deep comprehension; (\textit{\textbf{iii}})~\textbf{\textit{Video duration}}, ranging from $10$ seconds to over an hour and grouped as short, medium, or long, supporting evaluation of both short-term and long-range reasoning (See Fig.~\ref{fig:overview}b); and (\textbf{\textit{iv}})~\textit{\textbf{Difficulty level}}, categorized as easy, medium, or hard (See Fig.~\ref{fig:ablation2}b). 

The annotation process consists of three stages: video selection, question-answer annotation, and step-wise reasoning (see Fig.~\ref{fig:overview}c), requiring considerable expert effort. On average, $30$ mins to identify a suitable video, $40$ mins to craft a high-quality question-answer pair, and $1$ hour to compose detailed step-by-step reasoning, resulting in approximately $2$ to $2.5$ hours per sample. Across the dataset, this amounts to roughly $115$ person-days of annotation by science graduates. Different annotators handled each stage of a sample to ensure independent verification. Annotators could consult academic resources or LLMs for clarification.

\subsection{Video Selection}
\label{sec:video_selection}
We curate video sources that capture core challenges, including dynamic visual representations, long-range temporal dependencies, and the need for joint grounding across visual, textual, and audio modalities. We do this by selecting videos where the associated questions cannot be reliably answered using static frames or audio transcripts alone, instead focusing on content that exhibits rich temporal progression, such as step-by-step diagram construction and incremental equation derivation. Static slide presentations and videos with minimal visual transitions are excluded. All videos are sourced using the YouTube API, manually reviewed, and trimmed to retain only segments relevant to the annotated question. In addition, selection is guided by conceptual difficulty, ensuring coverage across a broad spectrum, from high-school topics to advanced university-level and Olympiad-style problems.

To construct the candidate video pool, we begin with a manually curated seed set and expand it using YouTube’s recommendation algorithm. After assembling a large set of video IDs, we apply automatic filtering based on metadata such as titles and descriptions. In the final stage, expert annotators manually review the videos review the videos and select those that meet our quality criteria and exhibit the desired reasoning types.
For chart-based questions, we include animated charts from documentary-style videos and public news articles, covering various chart types such as bar plots (vertical and horizontal), pie charts, line graphs, histograms and stacked bar charts. We specifically select videos that present multiple charts whose interpretations are temporally and conceptually linked. An example is shown in Fig.~\ref{fig:intro_fig}, where the first problem involves interpreting multiple related charts and reasoning across them to derive the answer. For the remaining nine mathematical categories, we collect a diverse mix of lecture videos, screen recordings, and digital whiteboard tutorials that reflect instructional formats commonly used in educational content.

\subsection{Question-Answer Annotation}
Each video in \benchmark is paired with a carefully constructed multiple-choice question designed to evaluate a model’s ability to reason over visual and temporal mathematical content. Questions are written to ensure that solving them requires meaningful engagement with the video, rather than relying on surface-level cues from transcripts or isolated frames. {This stage also serves as quality assurance for video selection, with approximately $70\%$ of selected videos retained, and the rest are discarded.} We categorize questions into three reasoning types: (i) \textit{Problem Focused}, where the question is explicitly stated and can be solved through direct observation and reasoning; (ii) \textit{Concept Transfer}, where a demonstrated method or solution approach must be applied to a similar but newly posed problem; and (iii) \textit{Deep Instructional Comprehension}, which involves following long instructional content, such as a lecture or tutorial, to understand the context, identify what is being asked, interpret the partially worked-out solution, and complete the solution. Fig.~\ref{fig:intro_fig} illustrates one example of each reasoning type, shown from left to right. All types require multi-step reasoning, with increasing levels of contextual and temporal integration across categories.

\subsection{Step-wise Reasoning and Quality Assessment}
Following question-answer annotation, each sample undergoes a second stage of annotation, where a separate annotator writes a \textit{step-by-step explanation} detailing the reasoning required to arrive at the final answer. Each question is assigned four to ten steps, with each step reflecting a meaningful semantic progression, capturing a distinct, essential part of the overall solution. This process results in a total of \textit{\textbf{2,945 high-quality expert-annotated reasoning steps}}. The steps are constructed in a logical sequential order, enabling step-level evaluation of how far a model progresses toward the correct answer; model-generated chain-of-thought responses are compared against these steps during evaluation. This stage also serves as quality assurance for the question-answer pairs, allowing the annotator to verify their correctness and fix any errors or improve clarity. Approximately $30\%$ of the questions were refined during this stage. Additionally, each question is tagged with a set of likely error categories to support structured error analysis during the step evaluation.

\subsection{Comparison with Existing Video Benchmarks} While recent video-language benchmarks offer broad coverage across domains, they primarily focus on perceptual or narrative understanding, often lacking the depth, structure, and multi-modal precision required for mathematical reasoning. In contrast, \benchmark~targets the `needle-in-a-multimodal-haystack' challenge by demanding tightly aligned interpretation across high-resolution visuals, spoken explanations, and textual content, characteristics underrepresented in prior works. Unlike benchmarks that emphasize short-form clips or general comprehension, our dataset centers on domain-specific, step-intensive reasoning with annotated chain-of-thought traces, enabling fine-grained evaluation of both intermediate and final responses. We present a comparison of \benchmark~ with recent benchmarks in terms of domain, duration, STEM focus, and step-wise CoT annotation in the Appendix Tab.~\ref{tab:benchmark_comparison}. By focusing on mathematical problem-solving in long instructional videos, it reveals challenges still underexplored in current video-language benchmarks.

\vspace{-1.0em}
\section{Experiments}
We evaluate a diverse set of models on \benchmark, including \boldmath\textbf{5}\unboldmath~\textit{\textbf{proprietary multimodal models}}: Claude-3.7-sonnet, GPT-4o, GPT-o4-mini, Gemini 2.0 Flash, and Gemini 1.5 Flash, and \boldmath\textbf{25}\unboldmath~\textit{\textbf{open-source models}} selected for their strong capabilities in video reasoning. We specifically focus on state-of-the-art video models, spanning \textit{\textbf{4~model size categories}}: 5B, 9B, 40B, and 80B parameters. The pool includes recent models with advanced reasoning abilities, including those supporting chain-of-thought prompting (Qwen2.5-VL~\cite{Qwen2.5-VL}, VideoChat-R1~\cite{videochatR1}, Video-R1~\cite{videoR1}). To contextualize the performance of video-specialized models, we include two baselines, one vision-blind LLM (Qwen-2.5), and one state-of-the-art image model (Qwen2.5-VL) evaluated on a single key frame.

\subsection{Evaluation Strategies}
\label{sec:evaluation_strategy}
\textbf{Inference Protocol:} Given that the benchmark emphasizes the importance of multimodal grounding and temporal reasoning across modalities, we choose a setting that ensures fairness by adapting input configurations to align with the strengths and operational capabilities of each model. Specifically, each model samples frames according to its optimal setting (LLaVA-OneVision~\cite{llavaov} samples 32 frames,  Qwen2.5-VL samples up to 768 frames, and Gemini has access to the full video). Correspondingly, subtitles are precisely aligned with the sampled frames, providing richer audio context to models capable of handling longer temporal sequences. This approach inherently rewards models equipped to reason over extended temporal context and to effectively integrate information across modalities. {LMMs-Eval}~\cite{lmms_eval2024, zhang2024lmmsevalrealitycheckevaluation} was used for our experiments and now includes official support for evaluating VideoMathQA.

We implement \textit{\textbf{four evaluation strategies}} to comprehensively assess model performance: \textbf{(\textit{i})~Multiple-Choice Evaluation (MCQ)}: Each question provides five answer options (one correct and four distractors). This direct evaluation format offers clear reproducibility without reliance on LLM-based scoring. However, it can inflate weak model scores, especially in smaller models (e.g., $\leq9B$). \textbf{(\textit{ii})~Multi-Binary Evaluation (MBin)}: To better distinguish performance in such cases, we construct binary-choice variants by pairing the correct answer against each distractor independently. A model must select the correct option across all pairs to be marked correct, significantly reducing randomness and more accurately revealing true model capabilities, particularly critical when evaluating across a wide range of model scales. \textbf{(\textit{iii})~Chain-of-Thought (CoT) vs. Direct Answering}: In direct evaluation, strict instruction-following is crucial since models are instructed to respond with only the final correct option, allowing immediate, format-based extraction without post-processing. In contrast, CoT evaluations encourage models to articulate detailed reasoning steps prior to answering, reflecting human-like problem-solving approaches, and providing leniency regarding response formatting. Here, we use a lightweight state-of-the-art open-source LLM (Qwen-3-4B) in non-thinking mode to extract the final answer. This setup allows us to analyze whether reasoning enhances performance. \textbf{(\textit{iv})~Step-wise Reasoning Evaluation}: For CoT responses, we further evaluate the quality of reasoning by comparing model-generated rationales with annotated solution steps (typically $4$-$10$ per question), each representing a distinct semantic progression. We use the Qwen-3-4B model in its thinking mode for this task, leveraging its strong mathematical reasoning ability to ensure fair and accurate evaluation. The model assigns each response a score between $0$ and $10$, along with a rationale. We further use the rationales to enable error analysis against predefined error categories, providing deep, actionable insights that highlights model limitations and reasoning gaps.

\textbf{Implementation Details:} For all models, we run direct answering for both MCQ ($420$ QA pairs) and MBin ($1680$ QA pairs). For models capable of reasoning, we additionally perform CoT evaluations for both MCQ and MBin. We use a light-weight LLM based post-processing to extract the option choice from the CoT responses to compute accuracy. Step-wise evaluation is conducted only for MCQ CoT responses. Using ground-truth steps, model response, and critique from step evaluation, we perform error analysis for selected models. We further ablate the effect of subtitles across all settings to analyze model sensitivity to multimodal input. All prompts used, including those for CoT prompting, postprocessing, step-wise evaluation, error analysis, and subtitle handling are detailed in Appendix~\ref{prompt:step-eval} to \ref{prompt:error-prompt}.

\subsection{Quantitative Analysis}

\definecolor{small}{RGB}{255, 255, 255} 
\definecolor{big}{RGB}{169, 204, 227} 

\begin{table*}[t]
\centering
\renewcommand{\arraystretch}{1.08}
\resizebox{\textwidth}{!}{%
\setlength{\tabcolsep}{4pt}
\begin{tabular}{l c | cc | cc | cccccccccc | ccc }
\toprule
\multirow{2}{*}{\textbf{Models}} & \multirow{2}{*}{\textbf{Size}} & 
\multicolumn{2}{c|}{\textbf{MCQ}} & 
\multicolumn{2}{c|}{\textbf{MBin}} & 
\multicolumn{10}{c|}{\textbf{Mathematic Concepts}} & 
\multicolumn{3}{c}{\textbf{Duration}} \\
\cmidrule(lr){3-6} \cmidrule(lr){7-16} \cmidrule(lr){17-19}
& & \textbf{V} & \textbf{+Sub} & \textbf{V} & \textbf{+Sub} & 
\textbf{GAng} & \textbf{GAre} & \textbf{GLen} & 
\textbf{Chart} & \textbf{Stat} & \textbf{Arth} & 
\textbf{Topo} & \textbf{Grph} & \textbf{Cntg} & \textbf{Pzle} & 
\textbf{Short} & \textbf{Med} & \textbf{Long} \\
\midrule 
Random & - & 17.4 & 17.4 & 7.9 & 7.9 & 8.7 & 7.0 & 7.8 & 10.7 & 8.7 & 3.9 & 6.7 & 11.1 & 2.6 & 11.1 & 9.0 & 7.8 & 6.9 \\
\midrule
\multicolumn{19}{c}{\textit{Proprietary Models}} \\
\midrule 
Claude-3.7-sonnet & - & \rgbrank{26.2}{50} & \rgbrank{27.1}{46} & \rgbrank{8.6}{0} & \rgbrank{9.5}{0} & 
\rgbrank{17.4}{43} & \rgbrank{9.9}{0} & \rgbrank{5.9}{0} & \rgbrank{8.0}{0} & 
\rgbrank{17.4}{25} & \rgbrank{11.5}{7} & \rgbrank{13.3}{25} & \rgbrank{5.6}{4} & 
\rgbrank{5.3}{0} & \rgbrank{9.3}{4} & \rgbrank{8.2}{0} & \rgbrank{11.0}{0} & 
\rgbrank{9.1}{4} \\

GPT-4o & - & \rgbrank{20.2}{0} & \rgbrank{24.5}{32} & \rgbrank{12.6}{4} & \rgbrank{13.6}{4} & 
\rgbrank{13.0}{25} & \rgbrank{12.7}{11} & \rgbrank{15.7}{4} & \rgbrank{12.0}{25} & 
\rgbrank{4.4}{0} & \rgbrank{17.3}{25} & \rgbrank{20.0}{46} & \rgbrank{5.6}{4} & 
\rgbrank{7.9}{4} & \rgbrank{20.4}{54} & \rgbrank{14.2}{4} & \rgbrank{15.6}{14} & 
\rgbrank{10.6}{11} \\

Gemini-1.5-Flash & - & \rgbrank{20.5}{4} & \rgbrank{23.1}{21} & \rgbrank{12.6}{4} & \rgbrank{17.6}{36} & 
\rgbrank{26.1}{61} & \rgbrank{15.5}{14} & \rgbrank{19.6}{29} & \rgbrank{9.3}{11} & 
\rgbrank{17.4}{25} & \rgbrank{23.1}{57} & \rgbrank{6.7}{0} & \rgbrank{22.2}{68} & 
\rgbrank{15.8}{46} & \rgbrank{24.1}{64} & \rgbrank{17.9}{25} & \rgbrank{22.1}{50} & 
\rgbrank{12.1}{29} \\

Gemini-2.0-Flash & - & \rgbrank{28.6}{71} & \rgbrank{31.7}{75} & \rgbrank{14.1}{14} & \rgbrank{20.5}{57} & 
\rgbrank{30.4}{75} & \rgbrank{23.9}{50} & \rgbrank{27.5}{64} & \rgbrank{13.3}{36} & 
\rgbrank{8.7}{7} & \rgbrank{19.2}{39} & \rgbrank{13.3}{25} & \rgbrank{16.7}{61} & 
\rgbrank{7.9}{4} & \rgbrank{33.3}{89} & \rgbrank{25.4}{57} & \rgbrank{24.0}{64} & 
\rgbrank{11.4}{25} \\
\midrule 
\multicolumn{19}{c}{\textit{Open-source Models ($<5$\textnormal{B})}} \\
\midrule 
Qwen2.5-VL~\cite{Qwen2.5-VL} & 3B & \rgbrank{26.9}{57} & \rgbrank{27.6}{50} & \rgbrank{19.3}{54} & \rgbrank{19.6}{56} & 
\rgbrank{26.1}{61} & \rgbrank{23.9}{50} & \rgbrank{23.5}{36} & \rgbrank{21.3}{93} & 
\rgbrank{34.8}{100} & \rgbrank{17.3}{25} & \rgbrank{26.7}{86} & \rgbrank{11.1}{32} & 
\rgbrank{15.8}{46} & \rgbrank{20.4}{54} & \rgbrank{25.4}{57} & \rgbrank{23.4}{61} & 
\rgbrank{15.9}{75} \\

InternVL2.5~\cite{internvl25} & 2B & \rgbrank{24.3}{36} & \rgbrank{20.7}{4} & \rgbrank{14.3}{18} & \rgbrank{14.5}{11} & 
\rgbrank{21.7}{54} & \rgbrank{9.9}{0} & \rgbrank{27.5}{64} & \rgbrank{10.7}{18} & 
\rgbrank{4.4}{0} & \rgbrank{15.4}{14} & \rgbrank{20.0}{46} & \rgbrank{0.0}{0} & 
\rgbrank{15.8}{46} & \rgbrank{16.7}{29} & \rgbrank{17.9}{25} & \rgbrank{16.9}{18} & 
\rgbrank{8.3}{0} \\

PLM-LLaMA~\cite{cho2025perceptionlm} & 3B & \rgbrank{22.9}{29} & \rgbrank{22.1}{14} & \rgbrank{13.6}{11} & \rgbrank{15.0}{18} & 
\rgbrank{17.4}{43} & \rgbrank{16.9}{25} & \rgbrank{25.5}{46} & \rgbrank{8.0}{0} & 
\rgbrank{26.1}{68} & \rgbrank{9.6}{0} & \rgbrank{20.0}{46} & \rgbrank{11.1}{32} & 
\rgbrank{13.2}{29} & \rgbrank{13.0}{21} & \rgbrank{16.4}{11} & \rgbrank{18.8}{29} & 
\rgbrank{9.1}{4} \\

InternVL3~\cite{zhu2025internvl3} & 2B & \rgbrank{22.4}{21} & \rgbrank{23.3}{29} & \rgbrank{18.8}{46} & \rgbrank{16.4}{29} & 
\rgbrank{21.7}{54} & \rgbrank{16.9}{25} & \rgbrank{17.7}{11} & \rgbrank{17.3}{64} & 
\rgbrank{30.4}{86} & \rgbrank{15.4}{14} & \rgbrank{20.0}{46} & \rgbrank{22.2}{68} & 
\rgbrank{13.2}{29} & \rgbrank{5.6}{0} & \rgbrank{18.7}{36} & \rgbrank{14.9}{7} & 
\rgbrank{15.9}{75} \\
\midrule 
\multicolumn{19}{c}{\textit{Open-source Models ($<9$\textnormal{B})}} \\
\midrule 
PLM-LLaMA~\cite{cho2025perceptionlm} & 8B & \rgbrank{22.1}{18} & \rgbrank{23.1}{21} & \rgbrank{16.7}{32} & \rgbrank{14.5}{11} & 
\rgbrank{13.0}{25} & \rgbrank{11.3}{7} & \rgbrank{17.7}{11} & \rgbrank{13.3}{36} & 
\rgbrank{17.4}{25} & \rgbrank{17.3}{25} & \rgbrank{20.0}{46} & \rgbrank{11.1}{32} & 
\rgbrank{10.5}{11} & \rgbrank{16.7}{29} & \rgbrank{16.4}{11} & \rgbrank{14.9}{7} & 
\rgbrank{12.1}{29} \\

Oryx-1.5~\cite{liu2024oryx} & 7B & \rgbrank{22.6}{25} & \rgbrank{22.6}{18} & \rgbrank{16.9}{36} & \rgbrank{17.4}{32} & 
\rgbrank{13.0}{25} & \rgbrank{23.9}{50} & \rgbrank{23.5}{36} & \rgbrank{9.3}{11} & 
\rgbrank{21.7}{54} & \rgbrank{23.1}{57} & \rgbrank{20.0}{46} & \rgbrank{5.6}{4} & 
\rgbrank{18.4}{75} & \rgbrank{11.1}{11} & \rgbrank{20.2}{43} & \rgbrank{20.8}{43} & 
\rgbrank{10.6}{11} \\

LLaVA-OV~\cite{llavaov} & 7B & \rgbrank{20.7}{7} & \rgbrank{21.2}{7} & \rgbrank{14.8}{21} & \rgbrank{15.5}{21} & 
\rgbrank{8.7}{0} & \rgbrank{15.5}{14} & \rgbrank{17.7}{11} & \rgbrank{16.0}{57} & 
\rgbrank{30.4}{86} & \rgbrank{17.3}{25} & \rgbrank{13.3}{25} & \rgbrank{5.6}{4} & 
\rgbrank{15.8}{46} & \rgbrank{11.1}{11} & \rgbrank{16.4}{11} & \rgbrank{18.8}{29} & 
\rgbrank{10.6}{11} \\

LongVA-DPO~\cite{longva} & 7B & \rgbrank{21.4}{11} & \rgbrank{21.7}{11} & \rgbrank{16.2}{29} & \rgbrank{14.1}{7} & 
\rgbrank{8.7}{0} & \rgbrank{15.5}{14} & \rgbrank{17.7}{11} & \rgbrank{12.0}{25} & 
\rgbrank{30.4}{86} & \rgbrank{9.6}{0} & \rgbrank{6.7}{0} & \rgbrank{5.6}{4} & 
\rgbrank{10.5}{11} & \rgbrank{18.5}{39} & \rgbrank{14.9}{7} & \rgbrank{11.7}{4} & 
\rgbrank{15.9}{75} \\

Video-R1~\cite{videoR1} & 7B & \rgbrank{21.4}{11} & \rgbrank{17.4}{0} & \rgbrank{16.0}{25} & \rgbrank{16.2}{25} & 
\rgbrank{8.7}{0} & \rgbrank{22.5}{39} & \rgbrank{25.5}{46} & \rgbrank{16.0}{57} & 
\rgbrank{26.1}{68} & \rgbrank{13.5}{11} & \rgbrank{6.7}{0} & \rgbrank{5.6}{4} & 
\rgbrank{13.2}{29} & \rgbrank{9.3}{4} & \rgbrank{16.4}{11} & \rgbrank{16.9}{18} & 
\rgbrank{15.2}{68} \\

InternVL2.5~\cite{internvl25} & 8B & \rgbrank{24.3}{39} & \rgbrank{24.8}{36} & \rgbrank{18.6}{43} & \rgbrank{18.6}{39} & 
\rgbrank{26.1}{61} & \rgbrank{19.7}{32} & \rgbrank{17.7}{11} & \rgbrank{17.3}{64} & 
\rgbrank{21.7}{54} & \rgbrank{19.2}{39} & \rgbrank{26.7}{86} & \rgbrank{11.1}{32} & 
\rgbrank{10.5}{11} & \rgbrank{20.4}{54} & \rgbrank{17.9}{25} & \rgbrank{22.7}{57} & 
\rgbrank{14.4}{57} \\

LLaVA-Video~\cite{llavavideo178} & 7B & \rgbrank{26.9}{57} & \rgbrank{26.4}{43} & \rgbrank{20.0}{61} & \rgbrank{19.3}{54} & 
\rgbrank{13.0}{25} & \rgbrank{21.1}{36} & \rgbrank{31.4}{75} & \rgbrank{17.3}{64} & 
\rgbrank{17.4}{25} & \rgbrank{15.4}{14} & \rgbrank{26.7}{86} & \rgbrank{5.6}{4} & 
\rgbrank{18.4}{75} & \rgbrank{18.5}{39} & \rgbrank{23.9}{54} & \rgbrank{20.8}{43} & 
\rgbrank{12.9}{39} \\

InternVideo2.5~\cite{internvideo25} & 8B & \rgbrank{25.2}{43} & \rgbrank{28.6}{64} & \rgbrank{19.1}{50} & \rgbrank{19.1}{43} & 
\rgbrank{34.8}{82} & \rgbrank{22.5}{39} & \rgbrank{15.7}{4} & \rgbrank{14.7}{50} & 
\rgbrank{21.7}{54} & \rgbrank{19.2}{39} & \rgbrank{20.0}{46} & \rgbrank{27.8}{82} & 
\rgbrank{10.5}{11} & \rgbrank{18.5}{39} & \rgbrank{18.7}{36} & \rgbrank{22.1}{50} & 
\rgbrank{15.9}{75} \\

Qwen2.5-VL~\cite{Qwen2.5-VL} & 7B & \rgbrank{26.7}{54} & \rgbrank{27.9}{54} & \rgbrank{19.8}{57} & \rgbrank{19.1}{43} & 
\rgbrank{8.7}{0} & \rgbrank{25.4}{61} & \rgbrank{25.5}{46} & \rgbrank{18.7}{79} & 
\rgbrank{13.0}{11} & \rgbrank{23.1}{57} & \rgbrank{13.3}{25} & \rgbrank{5.6}{4} & 
\rgbrank{15.8}{46} & \rgbrank{16.7}{29} & \rgbrank{22.4}{50} & \rgbrank{19.5}{36} & 
\rgbrank{15.2}{68} \\

InternVL3~\cite{zhu2025internvl3} & 8B & \rgbrank{29.1}{75} & \rgbrank{27.9}{54} & \rgbrank{20.0}{61} & \rgbrank{20.7}{61} & 
\rgbrank{13.0}{25} & \rgbrank{29.6}{68} & \rgbrank{27.5}{64} & \rgbrank{13.3}{36} & 
\rgbrank{13.0}{11} & \rgbrank{28.9}{75} & \rgbrank{20.0}{46} & \rgbrank{22.2}{68} & 
\rgbrank{15.8}{46} & \rgbrank{14.8}{25} & \rgbrank{25.4}{57} & \rgbrank{24.0}{64} & 
\rgbrank{12.1}{29} \\

VideoChat-R1~\cite{videochatR1} & 7B & \rgbrank{27.6}{64} & \rgbrank{29.1}{68} & \rgbrank{21.2}{75} & \rgbrank{21.2}{64} & 
\rgbrank{8.7}{0} & \rgbrank{22.5}{39} & \rgbrank{31.4}{75} & \rgbrank{21.3}{93} & 
\rgbrank{17.4}{25} & \rgbrank{30.8}{79} & \rgbrank{6.7}{0} & \rgbrank{11.1}{32} & 
\rgbrank{15.8}{46} & \rgbrank{18.5}{39} & \rgbrank{26.9}{68} & \rgbrank{20.1}{39} & 
\rgbrank{16.7}{89} \\
\midrule 
\multicolumn{19}{c}{\textit{Open-source Models ($<40$\textnormal{B})}} \\
\midrule 
Aria~\cite{li2024aria} & 34B & \rgbrank{23.8}{32} & \rgbrank{26.4}{39} & \rgbrank{17.4}{39} & \rgbrank{19.1}{43} & 
\rgbrank{8.7}{0} & \rgbrank{25.4}{61} & \rgbrank{19.6}{29} & \rgbrank{22.7}{100} & 
\rgbrank{17.4}{25} & \rgbrank{19.2}{39} & \rgbrank{20.0}{46} & \rgbrank{11.1}{32} & 
\rgbrank{21.1}{93} & \rgbrank{11.1}{11} & \rgbrank{21.6}{46} & \rgbrank{16.9}{18} & 
\rgbrank{18.9}{96} \\

Oryx-1.5~\cite{liu2024oryx} & 32B & \rgbrank{30.5}{79} & \rgbrank{33.1}{89} & \rgbrank{22.9}{79} & \rgbrank{24.1}{71} & 
\rgbrank{30.4}{75} & \rgbrank{39.4}{96} & \rgbrank{31.4}{75} & \rgbrank{10.7}{18} & 
\rgbrank{17.4}{25} & \rgbrank{21.2}{54} & \rgbrank{6.7}{0} & \rgbrank{11.1}{32} & 
\rgbrank{15.8}{46} & \rgbrank{33.3}{89} & \rgbrank{27.6}{71} & \rgbrank{29.9}{82} & 
\rgbrank{13.6}{46} \\

Qwen2.5-VL~\cite{Qwen2.5-VL} & 32B & \rgbrank{32.4}{89} & \rgbrank{32.6}{86} & \rgbrank{25.7}{89} & \rgbrank{24.8}{79} & 
\rgbrank{43.5}{96} & \rgbrank{31.0}{71} & \rgbrank{25.5}{46} & \rgbrank{14.7}{50} & 
\rgbrank{26.1}{68} & \rgbrank{26.9}{71} & \rgbrank{6.7}{0} & \rgbrank{27.8}{82} & 
\rgbrank{10.5}{11} & \rgbrank{33.3}{89} & \rgbrank{28.4}{79} & \rgbrank{30.5}{86} & 
\rgbrank{14.4}{57} \\

InternVL2.5~\cite{internvl25} & 38B & \rgbrank{31.0}{82} & \rgbrank{33.6}{93} & \rgbrank{24.1}{82} & \rgbrank{26.0}{86} & 
\rgbrank{43.5}{96} & \rgbrank{38.0}{93} & \rgbrank{39.2}{100} & \rgbrank{8.0}{0} & 
\rgbrank{13.0}{11} & \rgbrank{32.7}{82} & \rgbrank{6.7}{0} & \rgbrank{11.1}{32} & 
\rgbrank{18.4}{75} & \rgbrank{29.6}{79} & \rgbrank{34.3}{93} & \rgbrank{31.8}{89} & 
\rgbrank{10.6}{11} \\

InternVL3~\cite{zhu2025internvl3} & 38B & \rgbrank{31.7}{86} & \rgbrank{35.7}{96} & \rgbrank{25.2}{86} & \rgbrank{29.5}{100} & 
\rgbrank{34.8}{82} & \rgbrank{42.3}{100} & \rgbrank{37.3}{96} & \rgbrank{13.3}{36} & 
\rgbrank{17.4}{25} & \rgbrank{25.0}{68} & \rgbrank{13.3}{25} & \rgbrank{33.3}{100} & 
\rgbrank{26.3}{100} & \rgbrank{40.7}{100} & \rgbrank{35.8}{100} & \rgbrank{38.3}{100} & 
\rgbrank{12.9}{39} \\

\midrule 
\multicolumn{19}{c}{\textit{Open-source Models ($<80$\textnormal{B})}} \\
\midrule 
LLaVA-Video~\cite{llavavideo178} & 72B & 
\rgbrank{28.3}{68} & \rgbrank{30.0}{71} & \rgbrank{20.2}{68} & \rgbrank{24.3}{75} & 
\rgbrank{8.7}{0} & \rgbrank{32.4}{79} & \rgbrank{25.5}{46} & \rgbrank{20.0}{89} & 
\rgbrank{13.0}{11} & \rgbrank{36.5}{93} & \rgbrank{13.3}{25} & \rgbrank{22.2}{68} & 
\rgbrank{21.1}{93} & \rgbrank{24.1}{64} & \rgbrank{27.6}{71} & \rgbrank{27.3}{71} & 
\rgbrank{17.4}{93} \\

LLaVA-OV~\cite{llavaov} & 72B & 
\rgbrank{25.5}{46} & \rgbrank{28.3}{61} & \rgbrank{21.0}{71} & \rgbrank{24.8}{79} & 
\rgbrank{17.4}{43} & \rgbrank{31.0}{71} & \rgbrank{23.5}{36} & \rgbrank{12.0}{25} & 
\rgbrank{21.7}{54} & \rgbrank{38.5}{96} & \rgbrank{20.0}{46} & \rgbrank{27.8}{82} & 
\rgbrank{18.4}{75} & \rgbrank{31.5}{86} & \rgbrank{30.6}{82} & \rgbrank{28.6}{75} & 
\rgbrank{14.4}{57} \\

InternVL2.5~\cite{internvl25} & 78B & 
\rgbrank{33.3}{93} & \rgbrank{31.7}{79} & \rgbrank{28.3}{96} & \rgbrank{27.9}{89} & 
\rgbrank{39.1}{89} & \rgbrank{36.6}{82} & \rgbrank{31.4}{75} & \rgbrank{18.7}{79} & 
\rgbrank{26.1}{68} & \rgbrank{32.7}{82} & \rgbrank{26.7}{86} & \rgbrank{27.8}{82} & 
\rgbrank{13.2}{29} & \rgbrank{27.8}{71} & \rgbrank{33.6}{86} & \rgbrank{35.1}{93} & 
\rgbrank{13.6}{46} \\

Qwen2.5-VL~\cite{Qwen2.5-VL} & 72B & \rgbrank{36.9}{100} & \rgbrank{37.6}{100} & \rgbrank{26.0}{93} & \rgbrank{27.9}{89} & 
\rgbrank{26.1}{61} & \rgbrank{36.6}{82} & \rgbrank{31.4}{75} & \rgbrank{17.3}{64} & 
\rgbrank{30.4}{86} & \rgbrank{38.5}{96} & \rgbrank{20.0}{46} & \rgbrank{16.7}{61} & 
\rgbrank{18.4}{75} & \rgbrank{29.6}{79} & \rgbrank{34.3}{93} & \rgbrank{29.2}{79} & 
\rgbrank{19.7}{100} \\

InternVL3~\cite{zhu2025internvl3} & 78B & \rgbrank{33.3}{93} & \rgbrank{31.7}{79} & \rgbrank{28.3}{96} & \rgbrank{27.9}{89} & 
\rgbrank{39.1}{89} & \rgbrank{36.6}{82} & \rgbrank{31.4}{75} & \rgbrank{18.7}{79} & 
\rgbrank{26.1}{68} & \rgbrank{32.7}{82} & \rgbrank{26.7}{86} & \rgbrank{27.8}{82} & 
\rgbrank{13.2}{29} & \rgbrank{27.8}{71} & \rgbrank{33.6}{86} & \rgbrank{35.1}{93} & 
\rgbrank{13.6}{46} \\

\bottomrule
\end{tabular}%
}
\caption{Performance on \benchmark~ using \emph{direct answer prompting} in both MCQ and Multi-Binary (MBin) settings. We report results with video-only (V) and video+subtitle (+Sub) inputs to highlight the effect of subtitles on multimodal grounding. MBin performance is further broken down across ten mathematical concepts and three video duration categories. The table covers open-source models across multiple size tiers. The mathematic concepts are geometric angle (GAng), geometric area (GAre), geometric length (GLen), chart reading (Chart), statistics and probability (Stat), arithmetic and calculus (Arth), topology (Topo), graph theory (Grph), counting (Cntg) and puzzles (Pzle).
}
\label{tab:non_cot_table_2}
\vspace{-1.5em}
\end{table*}

\definecolor{small}{RGB}{255, 255, 255} 
\definecolor{big}{RGB}{169, 204, 227}

\begin{table*}[t]
\centering
\renewcommand{\arraystretch}{1.08}
\resizebox{\textwidth}{!}{%
\setlength{\tabcolsep}{4pt}
\begin{tabular}{l c | cc | cc | cccccccccc | ccc | c }
\toprule
\multirow{2}{*}{\textbf{Models}} & \multirow{2}{*}{\textbf{Size}} & 
\multicolumn{2}{c|}{\textbf{MCQ}} & 
\multicolumn{2}{c|}{\textbf{MBin}} & 
\multicolumn{10}{c|}{\textbf{Mathematic Concepts}} & 
\multicolumn{3}{c|}{\textbf{Duration}} &
\textbf{CoT} \\
\cmidrule(lr){3-4} \cmidrule(lr){5-6} \cmidrule(lr){7-16} \cmidrule(lr){17-19}
& & \textbf{V} & \textbf{+Sub} & \textbf{V} & \textbf{+Sub} & 
\textbf{GAng} & \textbf{GAre} & \textbf{GLen} & 
\textbf{Chart} & \textbf{Stat} & \textbf{Arth} & 
\textbf{Topo} & \textbf{Grph} & \textbf{Cntg} & \textbf{Pzle} & 
\textbf{Short} & \textbf{Med} & \textbf{Long} & \textbf{Eval} \\
\midrule 
Random & - & 17.4 & 17.4 & 7.9 & 7.9 & 8.7 & 7.0 & 7.8 & 10.7 & 8.7 & 3.9 & 6.7 & 11.1 & 2.6 & 11.1 & 9.0 & 7.8 & 6.9 & - \\
\midrule
\multicolumn{20}{l}{\textbf{\textit{Open-source ($<9$\textnormal{B})}}} \\
Video-R1~\cite{videoR1} & 7B & \rgbrank{23.8}{19} & \rgbrank{27.6}{19} & \rgbrank{18.1}{31} & \rgbrank{20.0}{44} & \rgbrank{13.0}{19} & \rgbrank{26.8}{50} & \rgbrank{23.5}{44} & \rgbrank{9.3}{6} & \rgbrank{13.0}{13} & \rgbrank{34.6}{81} & \rgbrank{20.0}{44} & \rgbrank{16.7}{19} & \rgbrank{18.4}{44} & \rgbrank{16.7}{0} & \rgbrank{21.6}{50} & \rgbrank{26.0}{44} & \rgbrank{11.4}{13} & \rgbrank{3.9}{50} \\
LLaVA-Video~\cite{llavavideo178} & 7B & \rgbrank{26.4}{44} & \rgbrank{23.6}{0} & \rgbrank{20.0}{56} & \rgbrank{16.0}{0} & \rgbrank{4.4}{0} & \rgbrank{15.5}{6} & \rgbrank{23.5}{44} & \rgbrank{16.0}{56} & \rgbrank{21.7}{38} & \rgbrank{7.7}{0} & \rgbrank{26.7}{88} & \rgbrank{0.0}{0} & \rgbrank{21.1}{75} & \rgbrank{18.5}{13} & \rgbrank{16.4}{6} & \rgbrank{16.9}{0} & \rgbrank{14.4}{25} & \rgbrank{2.7}{0} \\
Qwen2.5-VL~\cite{Qwen2.5-VL} & 7B & \rgbrank{25.2}{38} & \rgbrank{29.5}{44} & \rgbrank{17.6}{19} & \rgbrank{18.3}{13} & \rgbrank{13.0}{19} & \rgbrank{15.5}{6} & \rgbrank{11.8}{0} & \rgbrank{20.0}{88} & \rgbrank{21.7}{38} & \rgbrank{36.5}{88} & \rgbrank{13.3}{6} & \rgbrank{16.7}{19} & \rgbrank{10.5}{6} & \rgbrank{16.7}{0} & \rgbrank{16.4}{6} & \rgbrank{20.1}{6} & \rgbrank{18.2}{69} & \rgbrank{3.7}{38} \\
InternVL3~\cite{zhu2025internvl3} & 8B & \rgbrank{28.8}{56} & \rgbrank{26.9}{6} & \rgbrank{17.9}{25} & \rgbrank{20.0}{44} & \rgbrank{17.4}{38} & \rgbrank{22.5}{25} & \rgbrank{27.5}{63} & \rgbrank{13.3}{19} & \rgbrank{4.4}{0} & \rgbrank{17.3}{13} & \rgbrank{13.3}{6} & \rgbrank{16.7}{19} & \rgbrank{7.9}{0} & \rgbrank{24.1}{44} & \rgbrank{19.4}{31} & \rgbrank{23.4}{19} & \rgbrank{9.9}{6} & \rgbrank{3.4}{25} \\
InternVideo2.5~\cite{internvideo25} & 8B & \rgbrank{24.3}{25} & \rgbrank{27.6}{19} & \rgbrank{18.3}{38} & \rgbrank{19.8}{31} & \rgbrank{26.1}{44} & \rgbrank{25.4}{44} & \rgbrank{13.7}{6} & \rgbrank{14.7}{44} & \rgbrank{26.1}{69} & \rgbrank{21.2}{19} & \rgbrank{13.3}{6} & \rgbrank{27.8}{81} & \rgbrank{18.4}{44} & \rgbrank{18.5}{13} & \rgbrank{17.9}{25} & \rgbrank{25.3}{31} & \rgbrank{15.2}{38} & \rgbrank{3.0}{6} \\
VideoChat-R1~\cite{videochatR1} & 7B & \rgbrank{22.4}{0} & \rgbrank{28.3}{31} & \rgbrank{21.4}{63} & \rgbrank{19.8}{31} & \rgbrank{13.0}{19} & \rgbrank{29.6}{56} & \rgbrank{15.7}{13} & \rgbrank{10.7}{13} & \rgbrank{17.4}{19} & \rgbrank{26.9}{44} & \rgbrank{6.7}{0} & \rgbrank{22.2}{50} & \rgbrank{18.4}{44} & \rgbrank{24.1}{44} & \rgbrank{20.9}{44} & \rgbrank{25.3}{31} & \rgbrank{12.1}{19} & \rgbrank{3.6}{31} \\
\midrule

\multicolumn{20}{l}{\textbf{\textit{Open-source ($<40$\textnormal{B})}}} \\
Oryx-1.5~\cite{liu2024oryx} & 32B & \rgbrank{29.1}{63} & \rgbrank{33.6}{63} & \rgbrank{21.7}{69} & \rgbrank{25.2}{75} & \rgbrank{34.8}{63} & \rgbrank{35.2}{88} & \rgbrank{43.1}{94} & \rgbrank{13.3}{19} & \rgbrank{17.4}{19} & \rgbrank{21.2}{19} & \rgbrank{20.0}{44} & \rgbrank{22.2}{50} & \rgbrank{18.4}{44} & \rgbrank{22.2}{31} & \rgbrank{30.6}{94} & \rgbrank{29.2}{63} & \rgbrank{15.2}{38} & \rgbrank{3.7}{38} \\ 
InternVL3~\cite{zhu2025internvl3} & 38B & \rgbrank{30.0}{69} & \rgbrank{31.4}{56} & \rgbrank{21.7}{69} & \rgbrank{25.0}{69} & \rgbrank{43.5}{88} & \rgbrank{31.0}{75} & \rgbrank{25.5}{56} & \rgbrank{16.0}{56} & \rgbrank{17.4}{19} & \rgbrank{32.7}{63} & \rgbrank{20.0}{44} & \rgbrank{11.1}{6} & \rgbrank{13.2}{25} & \rgbrank{31.5}{81} & \rgbrank{26.9}{63} & \rgbrank{31.2}{75} & \rgbrank{15.9}{50} & \rgbrank{4.1}{56} \\
Qwen2.5-VL~\cite{Qwen2.5-VL} & 32B & \rgbrank{31.4}{75} & \rgbrank{36.9}{75} & \rgbrank{22.6}{81} & \rgbrank{27.1}{81} & \rgbrank{47.8}{100} & \rgbrank{29.6}{56} & \rgbrank{33.3}{81} & \rgbrank{16.0}{56} & \rgbrank{26.1}{69} & \rgbrank{32.7}{63} & \rgbrank{13.3}{6} & \rgbrank{16.7}{19} & \rgbrank{23.7}{88} & \rgbrank{29.6}{75} & \rgbrank{29.1}{88} & \rgbrank{32.5}{81} & \rgbrank{18.9}{88} & \rgbrank{4.9}{81} \\
\midrule 
\multicolumn{20}{l}{\textbf{\textit{Open-source ($<80$\textnormal{B})}}} \\
LLaVA-Video~\cite{llavavideo178} & 72B & \rgbrank{23.6}{13} & \rgbrank{29.3}{38} & \rgbrank{14.8}{13} & \rgbrank{18.6}{19} & \rgbrank{8.7}{6} & \rgbrank{22.5}{25} & \rgbrank{17.7}{19} & \rgbrank{14.7}{44} & \rgbrank{8.7}{6} & \rgbrank{21.2}{19} & \rgbrank{26.7}{88} & \rgbrank{11.1}{6} & \rgbrank{26.3}{100} & \rgbrank{20.4}{25} & \rgbrank{17.2}{19} & \rgbrank{21.4}{13} & \rgbrank{16.7}{56} & \rgbrank{3.1}{13} \\
LLaVA-OV~\cite{llavaov} & 72B & \rgbrank{23.3}{6} & \rgbrank{26.9}{6} & \rgbrank{14.3}{6} & \rgbrank{18.1}{6} & \rgbrank{8.7}{6} & \rgbrank{14.1}{0} & \rgbrank{19.6}{31} & \rgbrank{13.3}{19} & \rgbrank{21.7}{38} & \rgbrank{26.9}{44} & \rgbrank{20.0}{44} & \rgbrank{22.2}{50} & \rgbrank{10.5}{6} & \rgbrank{25.9}{56} & \rgbrank{15.7}{0} & \rgbrank{23.4}{19} & \rgbrank{14.4}{25} & \rgbrank{3.2}{19} \\
Qwen2.5-VL~\cite{Qwen2.5-VL} & 72B & \rgbrank{37.4}{94} & \rgbrank{36.9}{75} & \rgbrank{24.5}{88} & \rgbrank{28.6}{94} & \rgbrank{30.4}{56} & \rgbrank{31.0}{75} & \rgbrank{31.4}{75} & \rgbrank{24.0}{94} & \rgbrank{21.7}{38} & \rgbrank{50.0}{94} & \rgbrank{13.3}{6} & \rgbrank{22.2}{50} & \rgbrank{15.8}{38} & \rgbrank{25.9}{56} & \rgbrank{27.6}{69} & \rgbrank{34.4}{88} & \rgbrank{22.7}{94} & \rgbrank{5.0}{94} \\
InternVL3~\cite{zhu2025internvl3} & 78B & \rgbrank{34.1}{81} & \rgbrank{37.1}{88} & \rgbrank{25.2}{94} & \rgbrank{27.9}{88} & \rgbrank{39.1}{81} & \rgbrank{39.4}{94} & \rgbrank{33.3}{81} & \rgbrank{13.3}{19} & \rgbrank{26.1}{69} & \rgbrank{23.1}{38} & \rgbrank{33.3}{100} & \rgbrank{22.2}{50} & \rgbrank{10.5}{6} & \rgbrank{40.7}{100} & \rgbrank{28.4}{81} & \rgbrank{36.4}{94} & \rgbrank{17.4}{63} & \rgbrank{4.9}{81} \\
\midrule 
\multicolumn{20}{l}{\textbf{\textit{Proprietary Models}}} \\
Claude-3.7-sonnet & - & \rgbrank{24.8}{31} & \rgbrank{29.5}{50} & \rgbrank{12.1}{0} & \rgbrank{19.3}{25} & \rgbrank{34.8}{63} & \rgbrank{29.6}{56} & \rgbrank{19.6}{31} & \rgbrank{4.0}{0} & \rgbrank{26.1}{69} & \rgbrank{13.5}{6} & \rgbrank{20.0}{44} & \rgbrank{16.7}{19} & \rgbrank{21.1}{75} & \rgbrank{22.2}{31} & \rgbrank{23.1}{56} & \rgbrank{26.0}{44} & \rgbrank{7.6}{0} & \rgbrank{4.2}{63} \\
GPT-4o & - & \rgbrank{27.1}{50} & \rgbrank{34.3}{69} & \rgbrank{18.6}{44} & \rgbrank{22.9}{56} & \rgbrank{26.1}{44} & \rgbrank{22.5}{25} & \rgbrank{17.7}{19} & \rgbrank{17.3}{75} & \rgbrank{30.4}{94} & \rgbrank{32.7}{63} & \rgbrank{20.0}{44} & \rgbrank{33.3}{88} & \rgbrank{13.2}{25} & \rgbrank{25.9}{56} & \rgbrank{19.4}{31} & \rgbrank{29.9}{69} & \rgbrank{18.2}{69} & \rgbrank{4.9}{81} \\
Gemini-2.0-Flash & - & \rgbrank{35.2}{88} & \rgbrank{38.8}{94} & \rgbrank{19.5}{50} & \rgbrank{24.8}{63} & \rgbrank{34.8}{63} & \rgbrank{21.1}{19} & \rgbrank{27.5}{63} & \rgbrank{18.7}{81} & \rgbrank{21.7}{38} & \rgbrank{28.9}{56} & \rgbrank{13.3}{6} & \rgbrank{33.3}{88} & \rgbrank{18.4}{44} & \rgbrank{33.3}{94} & \rgbrank{27.6}{69} & \rgbrank{27.9}{56} & \rgbrank{18.2}{69} & \rgbrank{4.7}{69} \\
GPT-o4-mini & - & \rgbrank{49.8}{100} & \rgbrank{61.4}{100} & \rgbrank{42.1}{100} & \rgbrank{44.8}{100} & \rgbrank{43.5}{88} & \rgbrank{49.3}{100} & \rgbrank{45.1}{100} & \rgbrank{40.0}{100} & \rgbrank{65.2}{100} & \rgbrank{63.5}{100} & \rgbrank{20.0}{44} & \rgbrank{72.2}{100} & \rgbrank{23.7}{88} & \rgbrank{31.5}{81} & \rgbrank{45.5}{100} & \rgbrank{44.8}{100} & \rgbrank{42.4}{100} & \rgbrank{6.9}{100} \\
\bottomrule
\end{tabular}%
}
\caption{Performance on \benchmark~ using \emph{chain-of-thoughts prompting} in MCQ and Multi-Binary (MBin) settings. We report results with video-only (V) and video+subtitle (+Sub) inputs to highlight the effect of subtitles on multimodal grounding. MBin performance is further broken down across ten mathematical concepts and three video duration categories. The table covers open-source models across multiple size tiers.
}
\label{tab:cot_table_1}
\vspace{-1em}
\end{table*}

Tab.~\ref{tab:non_cot_table_2} presents the direct answer evaluation, and Tab.~\ref{tab:cot_table_1} shows the chain-of-thought (CoT) evaluation. Both tables cover the MCQ and MBin evaluation, with and without subtitles, providing a comprehensive view of model capabilities. Tab.~\ref{tab:cot_table_1} additionally reports CoT step evaluation scores based on alignment with annotated reasoning traces. Proprietary models tend to benefit significantly from CoT prompting, whereas open-source models show mixed gains. In step evaluation, GPT-o4-mini achieves the highest score of $6.9$, with Qwen2.5-VL-72B leading among open models with a score of $5.0$.

\textbf{How does model size influence performance?} Across both MCQ and MBin settings, we observe that \textit{model performance improves with scale}. We observe this trend in both evaluations with CoT prompting in Tab.~\ref{tab:cot_table_1} and direct answering in Tab.~\ref{tab:non_cot_table_2}. For instance, in the CoT MBin evaluation with subtitles, InternVL-3~\cite{zhu2025internvl3} shows consistent improvement across model scale: $20.0\%$ ($8$B) to $25.0\%$ ($38$B) and $27.9\%$ ($72$B). Comparable trends are observed for other video-MLLM models (LLaVA-Video~\cite{llavavideo178}, LLaVA-OneVision{~\cite{llavaov}, Qwen2.5-VL~\cite{Qwen2.5-VL}), indicating that larger models are better at retaining temporal context, focusing on key visual details, and grounding information across modalities, all crucial for video-based mathematical reasoning. \textit{However, scale alone does not determine performance. Smaller, newer models often outperform older, larger ones.} For example, InternVL-3-38B surpasses multiple $72$B models (LLaVA-Video-72B, LLaVA-OneVision-72B) in both CoT and direct answers. Newer models benefit from stronger architectures, improved visual understanding, and better reasoning, enabling them to outperform larger, previously SoTA models.

\begin{TakeawayBox}{Takeaways: Model Size and Performance}
Larger models typically perform better, but newer, smaller models often outperform older, larger ones, highlighting that architecture and training quality matter as much as scale.
\end{TakeawayBox}

\textbf{How do proprietary models compare to open-source models?} 
We evaluate five proprietary models with CoT prompting and find that Gemini-2.0-Flash and GPT-o4-mini deliver the best performance. GPT-o4-mini achieves the highest overall accuracy, with $44.8\%$ in CoT MBin with subtitles. It performs particularly well in complex reasoning categories such as arithmetic-calculus ($63.5\%$), statistics ($65.2\%$), and geometric area ($49.3\%$), with performance significantly higher than the average of other proprietary and open-source models. These results suggest that its strong performance stems from a better ability to integrate understanding across vision, audio, text, and background subject knowledge, enabling more coherent mathematical reasoning. While proprietary models continue to lead, our results show that \textit{the gap between proprietary and open-source models is narrowing}. Optimized open-source models such as Qwen2.5-VL-72B and InternVL-3-78B outperform several proprietary counterparts, including Claude-3.7-Sonnet, Gemini-2.0-Flash, and GPT-4o.

\begin{TakeawayBox}{Takeaways: Proprietary vs. Open-source Models}
The gap between proprietary and open-source models is narrowing, as optimized open-source architectures and enhanced multimodal integration now closely match or surpass proprietary models.
\end{TakeawayBox}

\textbf{How do subtitles influence model performance?} Subtitles consistently enhance model performance across both CoT evaluation (Tab.~\ref{tab:cot_table_1}) and direct answering (Tab.~\ref{tab:non_cot_table_2}), especially for larger open-source and proprietary models. However, the impact of subtitles is not uniform: smaller models (\textless5B and \textless9B) often show minimal or inconsistent gains. In contrast, reasoning-capable models like GPT-o4-mini improve from $42.1\%$ (video-only) to $44.8\%$ (video+sub), and Qwen2.5-VL improves from $24.5\%$ to $28.6\%$ in the CoT MBin setting. As shown in Fig.~\ref{fig:ablation_1}b, \textit{models with stronger reasoning capabilities benefit more from subtitles, while smaller models struggle to extract meaningful gains from them.} These improvements reflect the ability to integrate fine‑grained audio cues with visual frames, a \lq{}needle-in-a-multimodal-haystack\rq{} challenge, where critical information is distributed across modalities, and stronger reasoning models are better equipped to ground these disparate cues into coherent solutions, while others may overlook small but essential verbal cues.

\begin{TakeawayBox}{Takeaways: Subtitles and Multimodal Reasoning}
Stronger reasoning models gain more from subtitles by anchoring verbal cues to visual frames. This reflects a ‘needle-in-a-multimodal-haystack’ challenge, where key signals are scattered across modalities. In contrast, smaller models often fail to integrate these cues, missing critical information for reasoning.
\end{TakeawayBox}

\textbf{How does video length and frame sampling affect model performance?} We evaluate model performance across short (\textless$30$s), medium ($30$s–$2$min), and long ($2$min–$1$hr) video categories and observe two distinct trends (see Fig.~\ref{fig:ablation_1}a). \textit{First}, while most models perform reasonably well on short videos, accuracy typically improves on medium-length videos and declines for longer durations. These trends align with the three reasoning challenges targeted by the benchmark. Short videos often correspond to \textit{problem focused} questions, requiring the model to derive a solution, where success hinges on general mathematical competence and the ability to extract key visual or verbal cues. Medium-length videos commonly involve \textit{concept transfer} questions, where the model is first shown a solution or method, then asked to adapt it to a related problem, favoring models that can effectively comprehend the instructions. In contrast, long videos correspond to \textit{deep instructional comprehension} questions, which require following extended, often non-linear instructional sequences to interpret the context. Here, the informational load is higher, and cues essential for solving the problem may be distributed sparsely across modalities and time. This setting closely aligns with the central challenge, where overlooking even a few critical details can derail the entire reasoning process. \textit{Second}, we study how frame sampling influences performance by evaluating Qwen2.5-VL with $16$, $64$, $256$, and $768$-frame settings (see Fig.~\ref{fig:ablation_1}c). We find that \textit{increasing frame count provides consistent improvements, particularly for longer videos}: a $5$-point gain for short videos and up to $8$ points in long videos, highlighting that models capable of handling extended frame sequences and maintaining long-range temporal coherence are better equipped for video-based mathematical reasoning. Further, in Fig.~\ref{fig:ablation2}a, we compare video models with vision-blind text only and single-image models, highligting that in-depth temporal reasoning is required to perform well on our \benchmark~benchmark.

\begin{TakeawayBox}{Takeaways: Impact of Video Length and Reasoning Type}
Models perform relatively well on applying learned concepts, moderately on direct problem solving, and struggle most with following detailed instructions in longer videos. Performance declines on tasks requiring alignment of sparse, cross-modal cues over time, highlighting the challenge of sustaining attention and inference across complex multimodal sequences.
\end{TakeawayBox}

\begin{figure*}[!t]
  \centering
  \includegraphics[width=\textwidth]{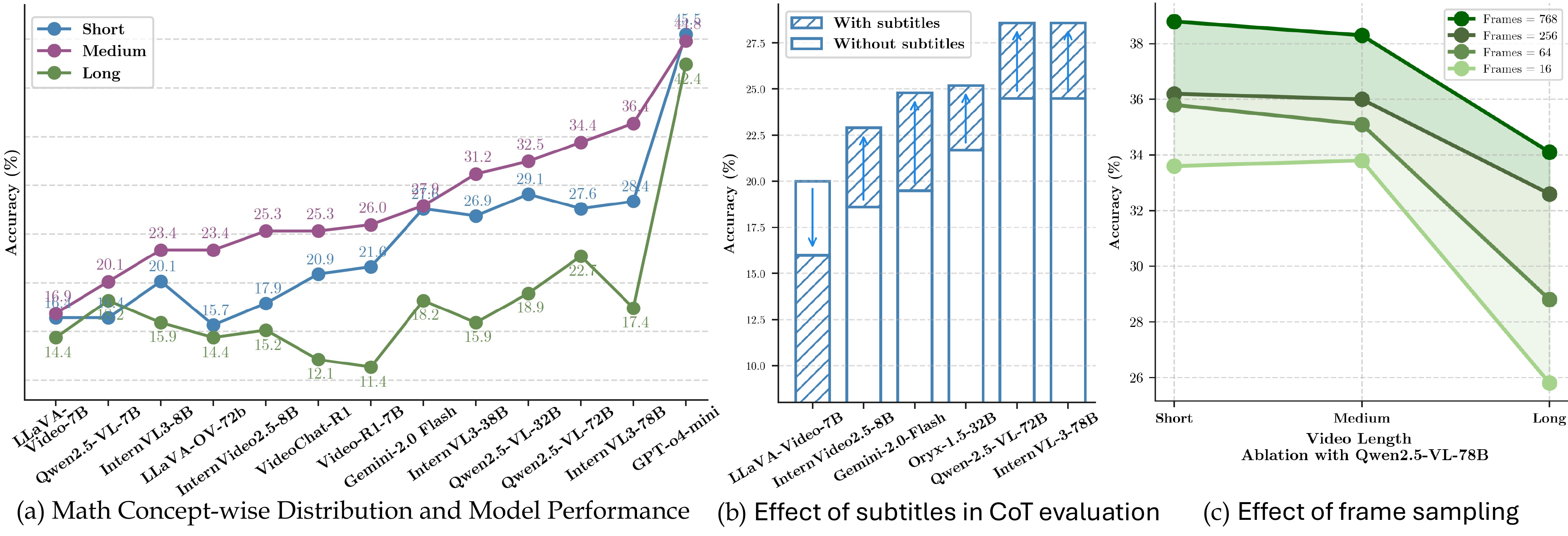}
  \vspace{-1em}
  \caption{The figure shows \benchmark~performance \textbf{a)} Across video duration categories using the CoT MBin +Sub setting; \textbf{b)} Impact of subtitles under the CoT MBin setting; and \textbf{c)} Effect of varying the number of input frames under CoT MCQ setting. Overall, models perform best on medium-length videos, and overall accuracy improves with the inclusion of subtitles and more frames during evaluation.}
  \label{fig:ablation_1}
  \vspace{-1em}
\end{figure*}

\textbf{How well do models perform across different mathematical concepts?} We analyze model performances across the ten mathematical categories covered in the benchmark and observe notable variation in their ability to comprehend and solve different mathematical concepts (see Fig.~\ref{fig:overview}a). Current models tend to perform better on questions involving \textit{arithmetic and calculus}, with average accuracy around $32$\% and GPT-o4-mini achieving the best performance of $63.5$\% with CoT evaluation. Most models show moderate performance on categories such as \textit{geometric reasoning and puzzles}, with average performance ranging between $24$\% and $30$\%. In contrast, \textit{chart reading, topology, graph theory, and statistics and probability} are more challenging for all models. Average accuracy across these categories typically falls between $16$\% and $21$\%, with GPT-o4-mini scoring only $20$\% in topology and graph theory, and a maximum of $40$\% in chart reading. 

\textbf{How does question difficulty affect model performance?} Model accuracy varies notably with question difficulty, while most models solve a moderate proportion of easy questions, they struggle with harder ones (Fig.~\ref{fig:ablation2}b). GPT-4o answers $96\%$ of easy questions correctly, yet handles only $46\%$ of the hard ones. InternVL-3-78B solves $60\%$ of easy examples, but manages just $8\%$ at the hard level. Other models follow the same trend, with limited success on more complex questions, exposing a key limitation in current models, their ability to generalize weakens under higher cognitive load.

\begin{figure*}[t]
  \centering
  \includegraphics[width=\textwidth]{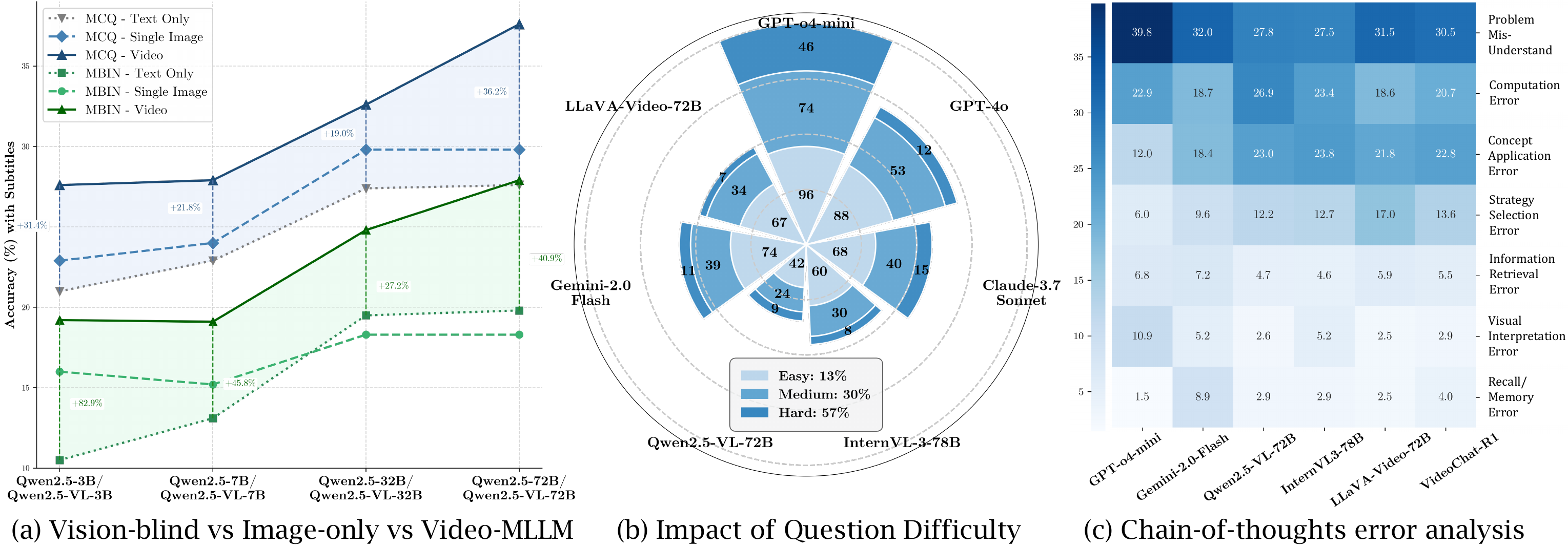}
  \caption{The figure shows \textbf{a)} Comparison among vision-blind, image-only, and video models, highlighting the need for video-level understanding to perform well in \benchmark. \textbf{b)} Distribution of questions in \benchmark across three difficulty levels for varying reasoning depths, and the relationship between performance and question difficulty across top-performing models. \textbf{c)} Error analysis based on CoT step evaluation. Most model errors stem from misunderstanding the question, where models misinterpret what the question asks or overlook critical multimodal cues.}
  \label{fig:ablation2}
\end{figure*}

\textbf{Where do models fail in the reasoning process?} Fig.~\ref{fig:ablation2}c presents a breakdown of model errors in the \benchmark based on CoT step evaluation across \textbf{\textit{seven error categories}}: misunderstanding the problem, failing to retrieve relevant information, misinterpreting visuals, incorrect application of concepts, choosing the wrong strategy or formula, forgetting previous context, and making calculation errors. Among these, the most common failure mode is problem misunderstanding, where models misinterpret what the question asks or overlook critical multimodal cues in the video. This reflects the core challenge of our benchmark, where missing even a small verbal or visual detail can derail reasoning entirely. \textit{Proprietary models} like GPT-o4-mini and Gemini-2.0-Flash \textit{show fewer errors in concept application and strategy selection} ($12$\% and $6$\% respectively), suggesting stronger domain grounding and better problem-solving execution. In contrast, open-source models like InternVL-3 exhibit more broadly distributed errors, with concept application and strategy selection together accounting for $23$\% of total errors, alongside notable mistakes in computation. Meanwhile, GPT-o4-mini shows a higher share of visual interpretation errors, indicating difficulty with fine-grained visual cues such as charts and diagrams. See Tab.~\ref{prompt:error-prompt} for details on the prompts used for error annotation and definitions of each error category.

\begin{TakeawayBox}{Takeaways: Common Failure Modes in Reasoning}
A key challenge for all models is extracting and integrating sparse information spread across modalities. Proprietary models show better integration of background subject knowledge, making fewer errors in concept application and strategy selection. Open-source models struggle more across reasoning steps, often applying the right idea incorrectly or choosing an ineffective path to the solution.
\end{TakeawayBox}

\section{Conclusion}
In this work, we introduce VideoMathQA, a comprehensive benchmark designed to evaluate mathematical reasoning in real-world educational video settings. By introducing carefully curated video-question pairs with detailed step-by-step reasoning trails, spanning diverse mathematical domains and instructional formats, our benchmark addresses the critical gap in the assessment of temporally extended, cross-modal reasoning capabilities in open-source and proprietary multimodal models. Unlike existing mathematical benchmarks, which mainly rely on static images and limited context, VideoMathQA requires models to jointly interpret dynamic visual, textual, and auditory information and to tackle real-world educational scenarios through problem solving, concept transfer, and deep instructional comprehension tasks.

\newpage
\appendix
 

\begin{center}

    {\huge \textbf{Appendix}}

\end{center}
 
\section{Comparison with Existing Benchmarks}
\begin{table*}[!h]
\centering
\resizebox{0.85\textwidth}{!}{%
\begin{tabular}{l|c|c|c|c|c}
\toprule
\textbf{Benchmark} & \textbf{Domain} & \textbf{Dur. (s)} & \textbf{STEM} & \textbf{CoT} & \textbf{Annotation} \\
\midrule
CLEVRER~\cite{clevrer}        & Physics (Syn.)    & 5       & \textcolor{darkgreen}{\textbf{\ding{51}}}   & \textcolor{red}{\textbf{\ding{55}}} & Auto \\
MovieChat-1K~\cite{moviechat}& Movie             & 500     & \textcolor{red}{\textbf{\ding{55}}} & \textcolor{red}{\textbf{\ding{55}}} & Human \\
EgoSchema~\cite{egoschema}    & Egocentric        & 180     & \textcolor{red}{\textbf{\ding{55}}} & \textcolor{red}{\textbf{\ding{55}}} & Auto+Human \\
Video-MME~\cite{videomme}     & General           & 1018    & \textcolor{red}{\textbf{\ding{55}}} & \textcolor{red}{\textbf{\ding{55}}} & Human \\
PerceptionTest~\cite{PerceptionTest}  & Perceptual  & 23      & \textcolor{red}{\textbf{\ding{55}}} & \textcolor{red}{\textbf{\ding{55}}} & Human \\
MMBench-Video~\cite{Mmbench}  & General           & $\sim$100 & \textcolor{red}{\textbf{\ding{55}}} & \textcolor{red}{\textbf{\ding{55}}} & Human \\
LongVideoBench~\cite{LongVideoBench}& General     & 473     & \textcolor{red}{\textbf{\ding{55}}} & \textcolor{red}{\textbf{\ding{55}}} & Human \\
Video-MMMU~\cite{Video-MMMU}  & Scientific        & 506     & \textcolor{darkgreen}{\textbf{\ding{51}}} & \textcolor{red}{\textbf{\ding{55}}} & Human \\
\midrule
\textbf{\benchmark (Ours)}    & Math              & 241 & \textcolor{darkgreen}{\textbf{\ding{51}}} & \textcolor{darkgreen}{\textbf{\ding{51}}} & \textbf{Human\dag} \\
\bottomrule
\end{tabular}
}
\vspace{0.5em}

\caption{Comparison of \benchmark~with recent video-language benchmarks in terms of domain, duration, STEM focus, step-wise CoT annotation, and source. \dag~Annotations for our \benchmark~are provided by graduate-level human experts.}
\label{tab:benchmark_comparison}
\end{table*}


\section{Additional Implementation Details}

We use $8~A100-80GB$ GPUs for all evaluations. For models with $\leq 8\text{B}$ parameters, we use data parallelism across 8 workers. For larger models, we utilize tensor parallelism with $TP=8$. All MLLM evaluations are conducted using lmms\_eval~\cite{lmms_eval2024}, while evaluations for language-only models are performed using VLLM~\cite{kwon2023efficient}. We provide the implementation of \benchmark\~in lmms\_eval, along with scripts to run both MLLM evaluations as part of the submission.

For all evaluated models, we use the recommended number of input frames following the official code base. Specifically, we use 128 frames for Aria~\cite{li2024aria}, 512 frames for InternVideo-2.5~\cite{internvideo25}, 64 frames for LLaVA-Video~\cite{llavavideo178}, 128 frames for LongVA~\cite{longva}, 32 frames for LLaVA-OneVision~\cite{llavaov} and 768 frames for Qwen2.5-VL~\cite{Qwen2.5-VL}. We use greedy decoding with a temperature of 0 for all MLLM evaluations. For chain of thought (CoT) evaluation, we post-process the responses to extract the final answer option using Qwen3-4B~\cite{qwen3} LLM. For the CoT step evaluation, we also use the Qwen3-4B model as the LLM judge and report the average over three runs. Below we provide all the LLM prompts we use in this work including CoT prompting, postprocessing, step-wise evaluation, error analysis and subtitle handling.

\section{Limitations and Future Work}
\benchmark~is an initial effort to select and annotate question-answer pairs along with step-by-step chain-of-thought reasoning in temporally rich videos, where the answer cannot be inferred merely from a few static frames or the audio transcript. The selection and annotation of these samples require a significant amount of time. For example, on average, it took graduate-level experts with at least a Master's degree in Science approximately 30 minutes to find a suitable video, 40 minutes to write a good question and answer, and 1 hour to compose detailed step-by-step reasoning. In total, annotating one sample took around 2 to 2.5 human hours, amounting to approximately 115 working days for 420 samples. This effort is substantial and makes scaling the dataset size difficult. We identify this as a limitation of our work and plan to explore semi-automatic annotation pipelines in the future to significantly reduce the annotation effort while maintaining high annotation quality.

\clearpage
\tcbset{
  promptbox/.style={
    colback=gray!5!white,
    colframe=black!75!white,
    fonttitle=\bfseries,
    title=LLM Evaluation Prompt for Chain of thoughts Step Evaluation,
    boxrule=0.4pt,
    arc=8pt,
    outer arc=8pt,
    top=4pt,
    bottom=4pt,
    left=4pt,
    right=4pt,
    enhanced,
  }
}

\phantomsection
\refstepcounter{subsection}
\label{prompt:step-eval}
\begin{tcolorbox}[promptbox]
You are a intelligent assistant for grading math question solutions. You will be given:
\begin{itemize}[itemsep=2pt, topsep=2pt, leftmargin=1em]
  \item A mathematical question (\texttt{question}) with multiple-choice options (\texttt{options}).
  \item A list of numbered ground truth steps (\texttt{gt\_steps}) showing the correct reasoning to solve a math problem.
  \item A answer (\texttt{answer}) that is the correct final solution to the question.
  \item A model prediction (\texttt{prediction}) that includes the steps the model followed and possibly the final answer.
\end{itemize}

\vspace{0.9em} 

\textbf{TASK:} Compare the prediction to gt\_steps and assign a score out of 10 using the rubric below. You must reward both matching logic and valid alternative reasoning. Avoid overly strict step-by-step comparison, instead focus if the model follows a coherent and plausible mathematical approach.

\vspace{0.9em} 

\textbf{Scoring Rubric:}
\begin{enumerate}[itemsep=6pt, topsep=2pt, leftmargin=1em]
  \item \textbf{Relative Step Matching (Main Criterion)}
  \begin{itemize}
    \item Count the total number of ground truth steps: N.
    \item Evaluate how many predicted steps correctly align with gt\_steps in terms of mathematical logic, reasoning, or computations.
    \item Score = (matching steps / N) × 10, rounded to nearest whole number.
    \item A step MATCHES if it serves the same mathematical purpose, even if phrased or ordered differently.
  \end{itemize}

  \item \textbf{Correct Final Answer via Different Reasoning}
  \begin{itemize}
    \item If the model's final answer is correct, and the steps are logically valid (even if they differ from \texttt{gt\_steps}, assign a full score of 10.
    \item Ignore number of matching steps in this case unless the reasoning is clearly flawed or incoherent.
    \item Reduce the score proportionally if reasoning contradicts parts of the ground truth.
  \end{itemize}

  \item \textbf{Implicit or Inferred Steps}
  \begin{itemize}
    \item Do NOT penalize if early steps are skipped, but later logic clearly depends on them.
    \item If a model does not state "identify the chart," but proceeds to use correct values, assume it did this implicitly.
    \item ALWAYS check for implied steps before reducing the score.
  \end{itemize}

  \item \textbf{Ignore Superficial Differences}
  \begin{itemize}
    \item Do NOT deduct score for formatting, different notation or variable names, or additional clarifications.
    \item FOCUS on the mathematical meaning, not the literal step match.
  \end{itemize}
\end{enumerate}

\vspace{0.9em} 

\textbf{Output Format}\\
\noindent\texttt{SCORE\_CARD: \{"matched\_steps": "X/N", "final\_answer\_correct": <0 or 1>, "critique": "<2--3 sentence summary>", "score": <0--10>\}}

\vspace{0.9em} 
Be strict when awarding credit. Do NOT be lenient. Carefully evaluate how far the model's reasoning aligns with the ground truth steps before assigning a score.
\end{tcolorbox}

\tcbset{
  promptbox/.style={
    colback=gray!5!white,
    colframe=black!75!white,
    fonttitle=\bfseries,
    title=LLM Evaluation Prompt for Error Analysis using CoT Step Evaluation,
    boxrule=0.4pt,
    arc=8pt,
    outer arc=8pt,
    top=4pt,
    bottom=4pt,
    left=4pt,
    right=4pt,
    enhanced,
  }
}
\phantomsection
\refstepcounter{subsection}
\begin{tcolorbox}[promptbox]
\label{prompt:error-prompt}
You are an intelligent assistant for analyzing model-generated math solutions. You will be given:
\begin{itemize}[itemsep=2pt, topsep=2pt, leftmargin=1em]
  \item A mathematical question (\texttt{question}) with multiple-choice options (\texttt{options}).
  \item A list of numbered ground truth steps (\texttt{gt\_steps}) showing the correct reasoning and a correct final solution (\texttt{answer}).
  \item A model prediction (\texttt{prediction}) that includes the steps the model followed.
  \item A critique (\texttt{critique}), which is a short rationale describing the quality of the model's reasoning and its alignment with the ground truth.
\end{itemize}

\vspace{0.9em} 

\textbf{TASK:} Your job is to carefully read and analyze the model's prediction.  Compare it to the ground truth steps and determine whether the prediction contains any of the following 7 types of reasoning errors. Use the critique and reason about where and why the model diverges from correct logic. Then, assign all relevant error category labels from the list below. The error types are not specific to each step; instead, identify a set of error types that apply to the overall reasoning.

\begin{itemize}[itemsep=2pt, topsep=2pt, leftmargin=1em]
  \item \textbf{TYPE1: Question Misunderstanding Error}: The model fails to understand what the question is asking or misunderstands its demand. It cannot correctly interpret which quantity, relationship, or part of the video is referenced, often mixing up figures, charts, or examples.
  \item \textbf{TYPE2: Information Retrieval Failure}: The model cannot locate or recognize the needed data in the video, including charts. It overlooks presented numbers, labels, angles, side lengths, diagrams, or text overlays, so no raw facts are available for further processing.
  \item \textbf{TYPE3: Visual Interpretation Error}: Although the model attends to the correct visual element, it reads it wrongly, misinterpreting axis scales, bar heights, marker positions, legends, or estimating distances and angles improperly.
  \item \textbf{TYPE4: Concept Application Error} The model knows which principle or method applies but is not able to execute it properly on the given question. It may recall the right concept yet misalign variables, swap parameters, or break the logical steps needed for that specific problem.
  \item \textbf{TYPE5: Strategy \& Formula Selection Error}: The model picks an entirely inappropriate approach, choosing the wrong theorem, formula, or problem‑solving strategy for the task.
  \item \textbf{TYPE6: Recall \& Memory Error}: The model forgets or ignores earlier information that are essential. It also covers cases when it contradicts key information from earlier in the question or in its own reasoning. This includes dropping previously used values, repeating steps unnecessarily, or breaking logical flow by not following through on earlier steps.
  \item \textbf{TYPE7: Computational Error}: The model has the correct inputs and method but makes calculation mistakes, incorrect addition, subtraction, multiplication, division, rounding, or unit‑conversion errors.
\end{itemize}

\vspace{0.7em} 
\textbf{RULES:}
\begin{itemize}[itemsep=2pt, topsep=2pt, leftmargin=1em]
  \item Multiple error types may apply to a single prediction. Errors may be global (relevant to all steps) or local (relevant to one or a few steps).
  \item Important Note: If the model uses different but valid reasoning and arrives at the correct answer, assign: none, even if the steps do not match the ground truth. 
  \item Do not assign an error unless you are confident it reflects a real mistake.
  \item If a clear mistake exists, but it does not match any of the listed error types, assign: uncategorized.
  \item Important Note: If you are not confident about which error applies, do not guess. Use uncategorized instead of forcing a type.
\end{itemize}

\vspace{0.7em}
\textbf{OUTPUT FORMAT:} Return only a comma-separated list of error type labels. Examples:
\begin{itemize}[itemsep=2pt, topsep=2pt, leftmargin=1em]
  \item \texttt{TYPE2, TYPE4}
  \item \texttt{none}
  \item \texttt{uncategorized}
\end{itemize}

\vspace{0.5em}
Do not include any explanations or extra text.
\end{tcolorbox}
\tcbset{
  cotpost/.style={
    colback=gray!5!white,
    colframe=black!75!white,
    fonttitle=\bfseries,
    title=Prompt for LLM Based Post Processing Multiple-Choice Chain-of-Thought Responses,
    boxrule=0.4pt,
    arc=8pt,
    outer arc=8pt,
    top=4pt,
    bottom=4pt,
    left=4pt,
    right=4pt,
    enhanced,
  }
}

\phantomsection
\refstepcounter{subsection}
\label{prompt:post_process}
\begin{tcolorbox}[cotpost]
Given the original multiple-choice options and a model-generated answer containing reasoning and a final answer, identify the option that best matches the final answer and return only the corresponding letter (A, B, C, D, or E).\\

The options are: \{\texttt{Options}\}\\

The model response is: \{\texttt{Response}\}\\

Only return the letter A, B, C, D, or E. If none is found, return \texttt{None}.
\end{tcolorbox}

\tcbset{
  mcqdirect/.style={
    colback=gray!5!white,
    colframe=black!75!white,
    fonttitle=\bfseries,
    title=Prompt for Multiple-Choice Evaluation with Direct Answering,
    boxrule=0.4pt,
    arc=8pt,
    outer arc=8pt,
    top=4pt,
    bottom=4pt,
    left=4pt,
    right=4pt,
    enhanced,
  }
}
\phantomsection
\refstepcounter{subsection}
\label{prompt:mcqdirect}
\begin{tcolorbox}[mcqdirect]
Select the best answer to the following multiple-choice question based on the video. Respond with only the letter (A, B, C, D or E) of the correct option.\\

Answer with the option's letter (A or B) from the given choices directly.
\end{tcolorbox}

\tcbset{
  mcqcot/.style={
    colback=gray!5!white,
    colframe=black!75!white,
    fonttitle=\bfseries,
    title=Prompt for Multiple-Choice Evaluation with Chain-of-Thought,
    boxrule=0.4pt,
    arc=8pt,
    outer arc=8pt,
    top=4pt,
    bottom=4pt,
    left=4pt,
    right=4pt,
    enhanced,
  }
}
\phantomsection
\refstepcounter{subsection}
\label{prompt:mcqcot}
\begin{tcolorbox}[mcqcot]
Select the best answer to the following multiple-choice question based on the video. Respond with the letter (A, B, C, D or E) of the correct option.\\

First, please perform reasoning, and think step-by-step to provide the best answer to the following question with the option's letter (A, B, C, D or E) from the given choices.
\end{tcolorbox}

\tcbset{
  mcqsubs/.style={
    colback=gray!5!white,
    colframe=black!75!white,
    fonttitle=\bfseries,
    title=Prompt for Multiple-Choice Evaluation with Subtitles,
    boxrule=0.4pt,
    arc=8pt,
    outer arc=8pt,
    top=4pt,
    bottom=4pt,
    left=4pt,
    right=4pt,
    enhanced,
  }
}
\phantomsection
\refstepcounter{subsection}
\label{prompt:mcqsubs}
\begin{tcolorbox}[mcqsubs]
The subtitles of the video are listed below: \\

\{\texttt{Subtitles}\}\\

Select the best answer to the following multiple-choice question based on the video. Respond with only the letter (A, B, C, D or E) of the correct option.
\end{tcolorbox}

\tcbset{
  mbindirect/.style={
    colback=gray!5!white,
    colframe=black!75!white,
    fonttitle=\bfseries,
    title=Prompt for Multi-Binary Evaluation with Direct Answering,
    boxrule=0.4pt,
    arc=8pt,
    outer arc=8pt,
    top=4pt,
    bottom=4pt,
    left=4pt,
    right=4pt,
    enhanced,
  }
}
\phantomsection
\refstepcounter{subsection}
\label{prompt:mbindirect}
\begin{tcolorbox}[mbindirect]
Select the best answer to the following multiple-choice question based on the video. Respond with only the letter (A or B) of the correct option.\\

Answer with the option's letter (A or B) from the given choices directly.
\end{tcolorbox}

\newpage
\bibliographystyle{lmrm}
\bibliography{main}

\end{document}